\documentclass[acmsmall]{acmart}
\AtBeginDocument{%
  \providecommand\BibTeX{{%
    \normalfont B\kern-0.5em{\scshape i\kern-0.25em b}\kern-0.8em\TeX}}}

\setcopyright{acmcopyright}
\copyrightyear{2023}
\acmYear{2023}
\acmDOI{XXXXXXX.XXXXXXX}

\usepackage{subcaption}
\usepackage{amsfonts}
\usepackage{amsmath,bm}
\usepackage{wrapfig}

\begin{document}


\title{Multi-View Graph Representation Learning Beyond Homophily}


\author{Bei Lin$^+$}
\email{lboctoberthirty@gmail.com}
\orcid{1234-5678-9012}

\author{You Li$^+$}
\email{youli.syvail@gmail.com}
\orcid{1234-5678-9012}

\author{Ning Gui}
\authornote{Corresponding Author.}
\email{ninggui@csu.edu.cn}

\author{Zhuopeng Xu}
\email{415793113@qq.com}
\affiliation{%
  \institution{School of Computer Science \& Engineering, Central South University}
  \streetaddress{YueLu Street}
  \city{Changsha}
  \state{Hunan}
  \country{China}
  \postcode{410083}}

\author{Zhiwu Yu}
\email{zhwyu@csu.edu.cn}
\affiliation{
\institution{National Engineering Research Center of High-speed Railway Construction Technology}
\city{Changsha}
\country{China}}



\renewcommand{\shortauthors}{Lin, Li, and Gui, et al.}

\begin{abstract}

Unsupervised graph representation learning(GRL) aims to distill diverse graph information into task-agnostic embeddings without label supervision. Due to a lack of support from labels, recent representation learning methods usually adopt self-supervised learning, and embeddings are learned by solving a handcrafted auxiliary task(so-called pretext task). However, partially due to the irregular non-Euclidean data in graphs, the pretext tasks are generally designed under homophily assumptions and cornered in the low-frequency signals, which results in significant loss of other signals, especially high-frequency signals widespread in graphs with heterophily. Motivated by this limitation, we propose a multi-view perspective and the usage of diverse pretext tasks to capture different signals in graphs into embeddings. A novel framework, denoted as Multi-view Graph Encoder(MVGE), is proposed, and a set of key designs are identified. More specifically, a set of new pretext tasks are designed to encode different types of signals, and a straightforward operation is propxwosed to maintain both the commodity and personalization in both the attribute and the structural levels. Extensive experiments on synthetic and real-world network datasets show that the node representations learned with MVGE achieve significant performance improvements in three different downstream tasks, especially on graphs with heterophily. Source code is available at \url{https://github.com/G-AILab/MVGE}.

\end{abstract}




\maketitle

\section{Introduction}

Graphs provide a natural and efficient representation for non-Euclidean data, such as brain networks, social networks, traffic maps in logistics\cite{lu2021dual}, and financial networks\cite{bronstein2017geometric}. To simplify the learning process, the current graph learning community generally divides graphs into two significant types from the node label's perspective: homophily and heterophily. In graphs with homophily, linked nodes often belong to the same class or have similar features, which is a crucial principle of many real-world networks, e.g., in the citation network, papers tend to cite papers from the same research area\cite{newman2018networks}. Partially due to its simplicity, this type of network has been under intensive study. In heterophily settings, trends in connections are reversed with the so-called "opposites attract" phenomenon. Linked nodes are likely from different classes or have different features, e.g., the chemical interactions in proteins often occur between different types of amino acids\cite{bo2021beyond}. Learning on graphs with heterophily is more difficult \cite{jia2020outcome}.

\begin{figure}[thbp]
\centering
	\begin{subfigure}{0.24\textwidth}
		\centering
		\includegraphics[width=\textwidth]{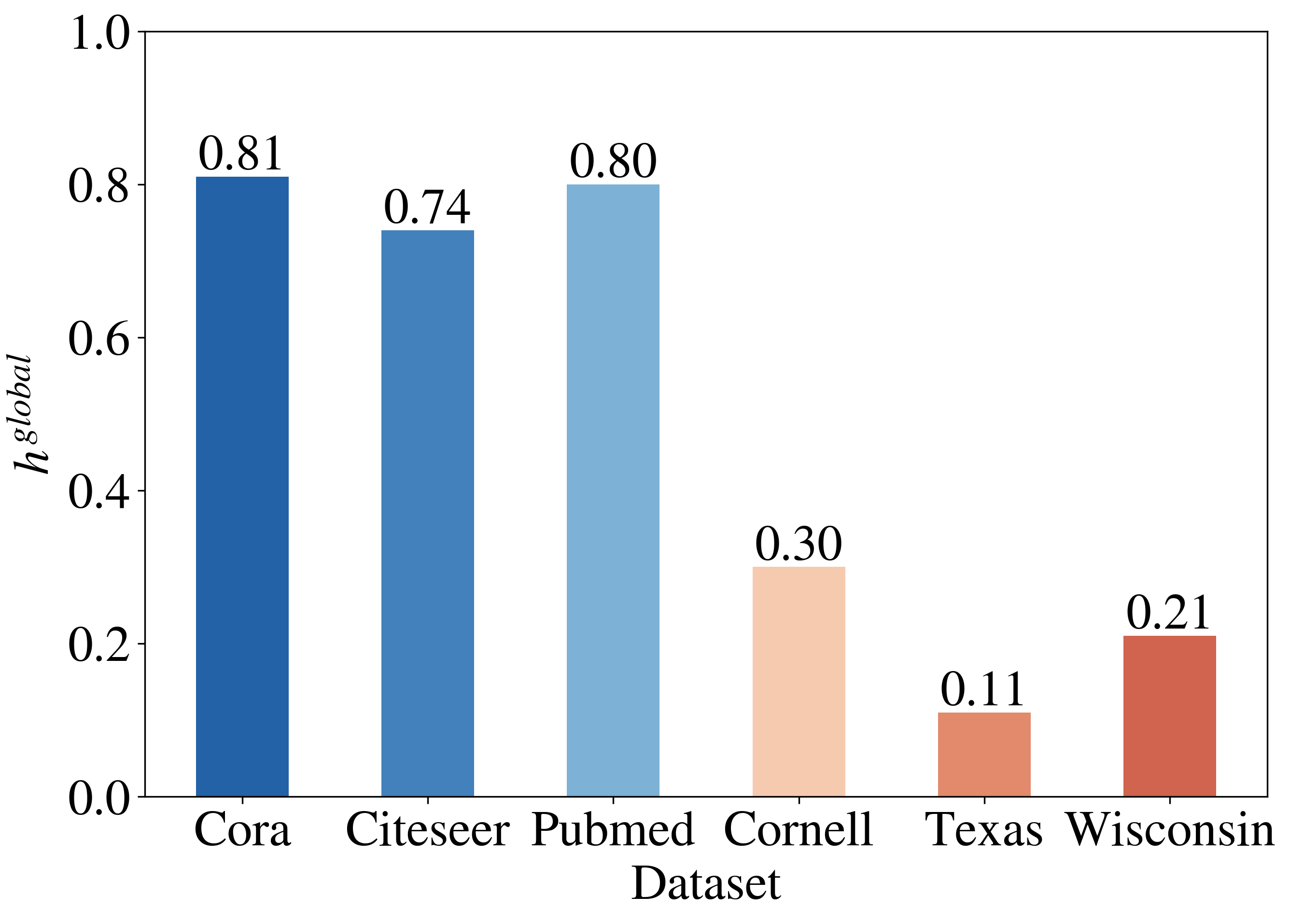}
		\caption{Global Homophily}
		\label{fig:GlobalHomophily}
	\end{subfigure}
	\begin{subfigure}{0.24\textwidth}
		\centering
		\includegraphics[width=\textwidth]{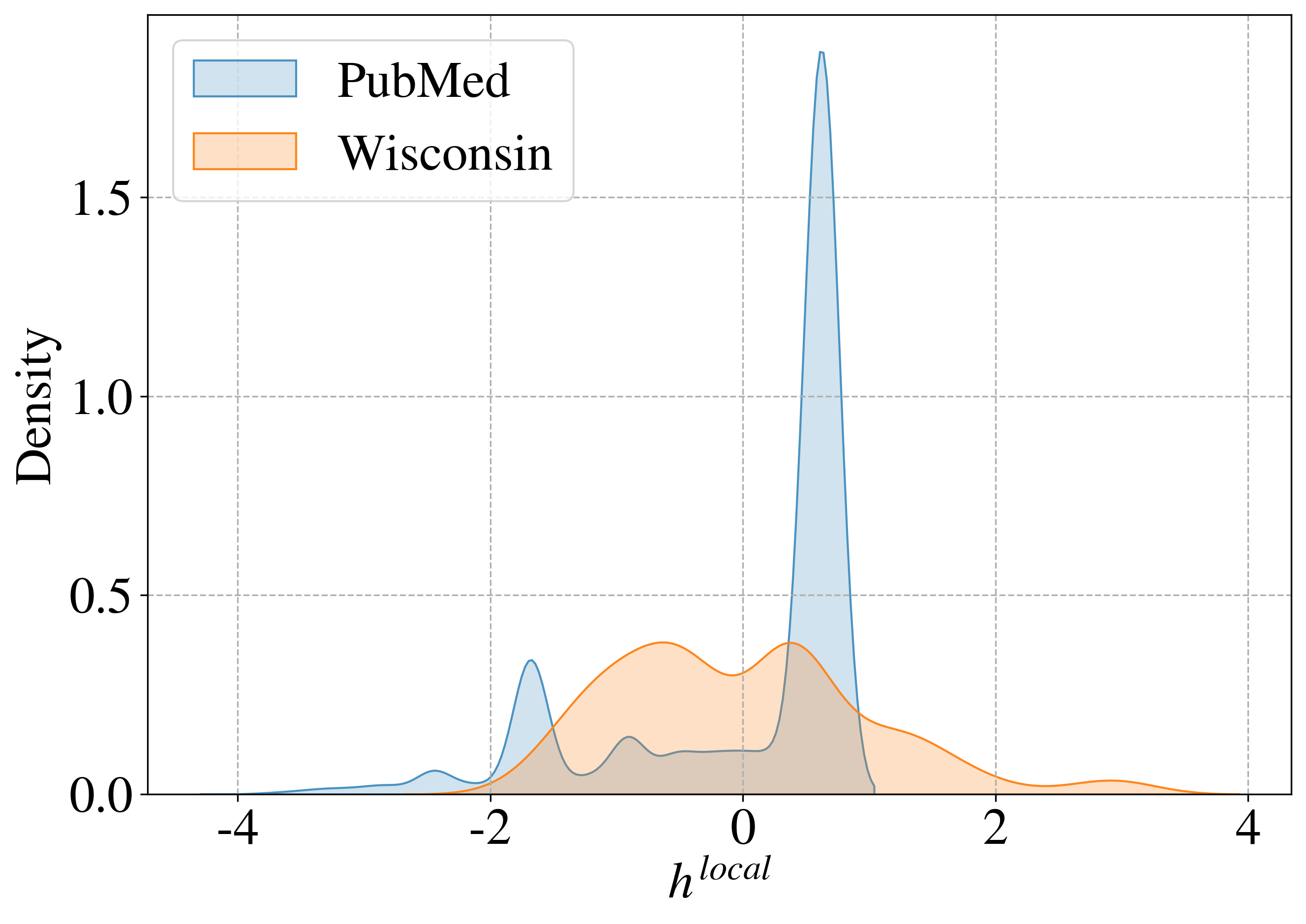}
		\caption{Local Homophily}
		\label{fig:LocalHomophily}
	\end{subfigure}
		\begin{subfigure}{0.24\textwidth}
		\centering
		\includegraphics[width=\textwidth]{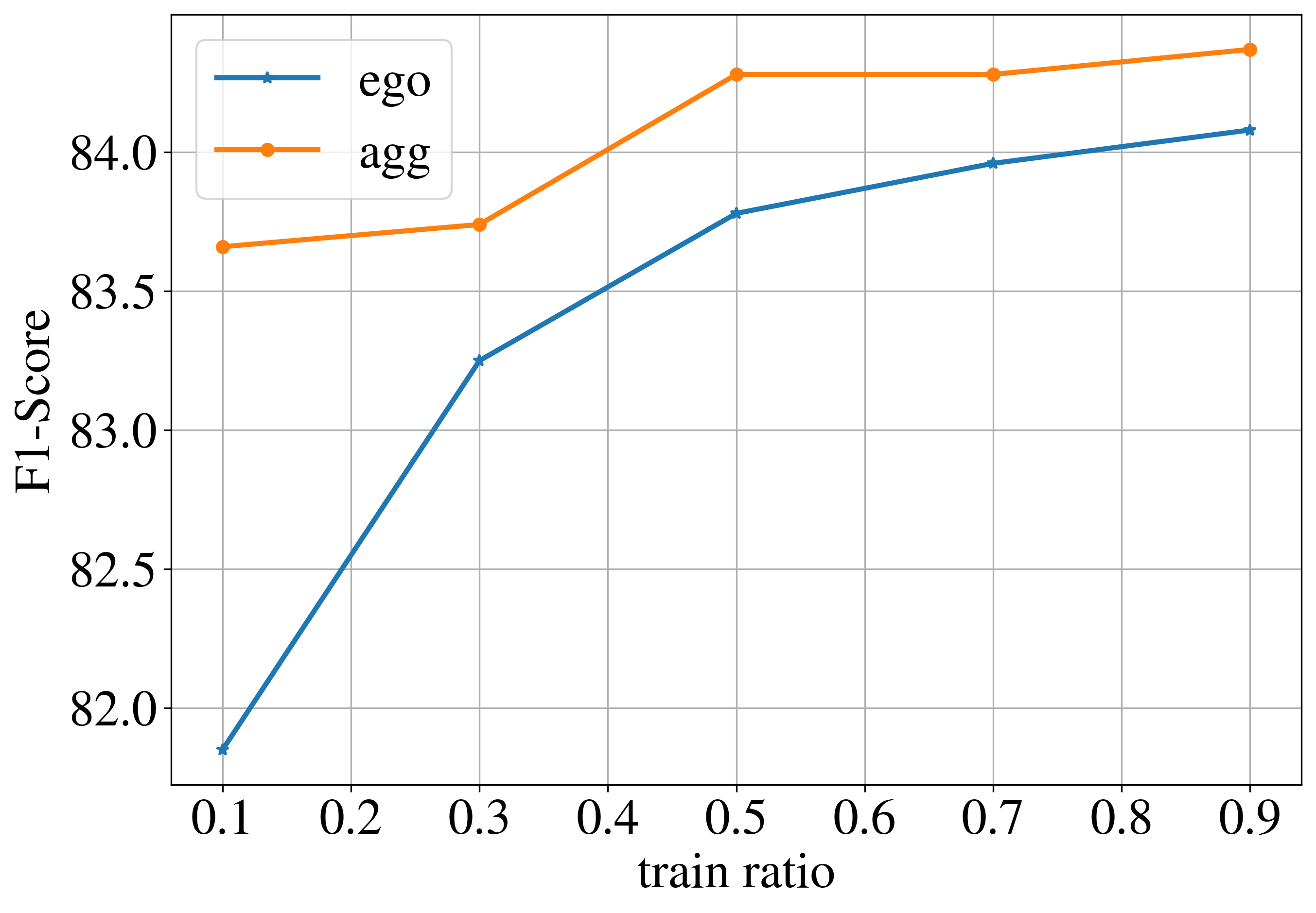}
		\caption{Node on Pubmed}
		\label{fig:High}
	\end{subfigure}
	\begin{subfigure}{0.24\textwidth}
		\centering
		\includegraphics[width=\textwidth]{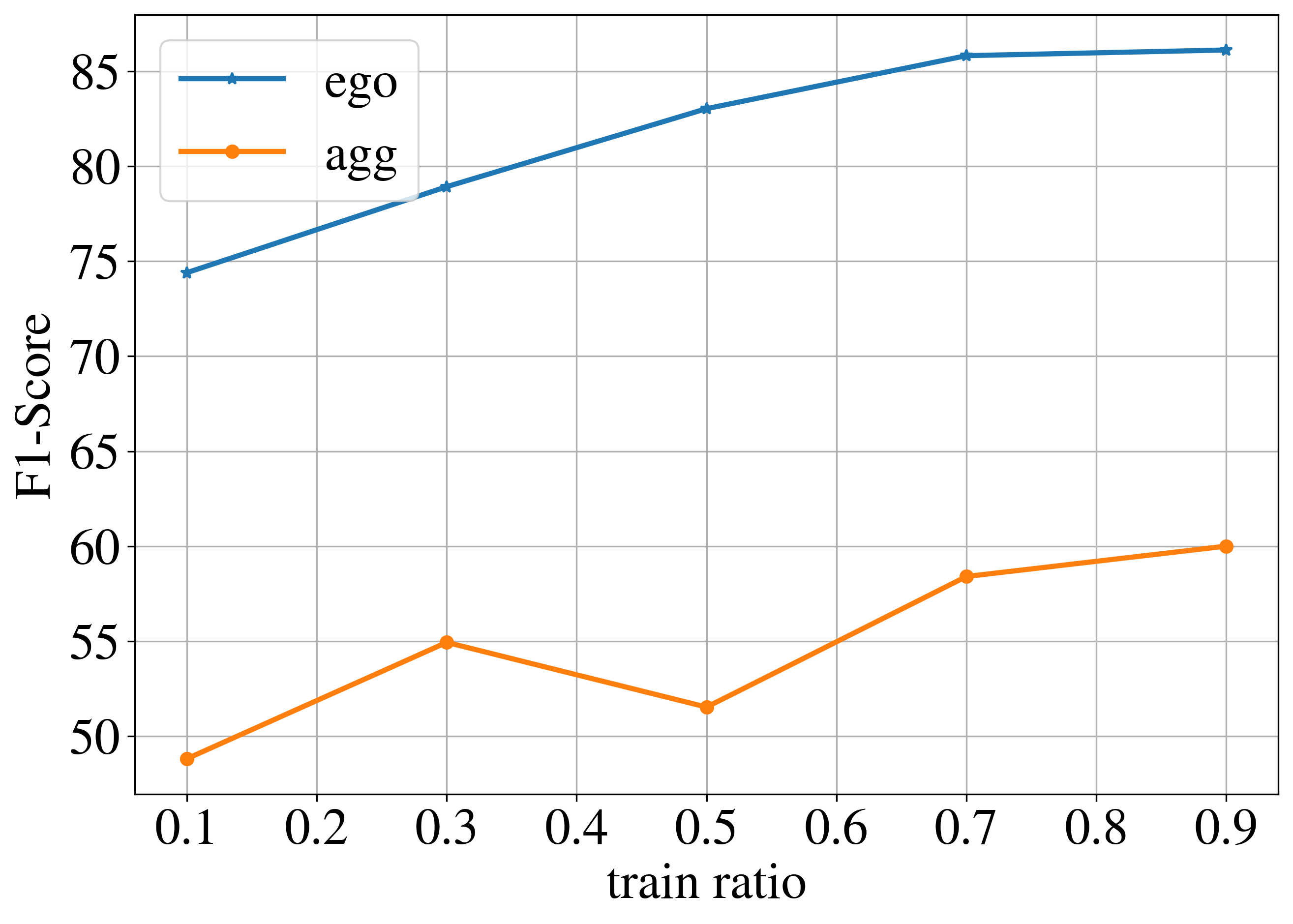}
		\caption{Node on Wisco.}
		\label{fig:low}
	\end{subfigure}
	\caption{(a) Global homophily of different real-world datasets. (b) Local homophily of Pubmed and Wisconsin. (c)-(d) Node Classification accuracy under homophily/heterophily settings. 'ego' and 'agg' denote ego feature reconstruction and aggregated feature reconstruction respectively.}
   \label{fig:discussion}
\end{figure}

However, either homophily or heterophily is classified based on the global assortativity \cite{newman2003mixing}, a metric defined only concerning node labels across the whole graph. In reality, real-world graphs can not be readily classified into either homophily or heterophily while they essentially reveal a complex and typically mixture of patterns from both node and attribute levels~\cite{peel2018multiscale,li2022graph}. Fig.~\ref{fig:GlobalHomophily} shows that real-world graphs have different degrees of homophily: Cora, Citeseer, and Pubmed are with high homophily, while Cornell, Texas, and Wisconsin are with high heterophily. Furthermore, even within a graph, the actual distribution patterns are highly diverse, Fig.~\ref{fig:LocalHomophily} shows that Cora with strong homophily still exist nodes with local heterophily, and vice versa also holds for Wisconsin with strong global heterophily. This fact clearly shows abundant signals, including low-frequency signals for commonality and high-frequency signals for disparity and the mixtures of those signals.

The goal of unsupervised Graph Representation Learning(GRL) is to condense different types of information existing in both the graph attributes and the topology into task-agnostic low-dimension embeddings \cite{hamilton2017representation}. 
Such complex patterns in high-dimensional and intractable non-Euclidean graph data bring enormous challenges for unsupervised GRL research without direct access to ground truth labels during embeddings. Due to the lack of label guidelines and the popularity of homophily, most GRL solutions are designed under strong homophily assumption\cite{li2022graph}. For example, random walk-based methods(e.g., DeepWalk\cite{perozzi2014deepwalk}, Node2Vec\cite{grover2016node2vec}) match nodes’ co-occurrence rate on short random walks over the graph to force neighboring nodes to have similar representations. GAE, VGAE\cite{kipf2016variational} and later variants ARVGE\cite{pan2018adversarially} assumes the connected nodes have more similar embeddings. GNN-based solutions are optimized for the network with homophily by propagating features and aggregating them within various graph neighborhoods via different mechanisms (e.g., averaging, LSTM, self-attention)\cite{zhu2020beyond}, e.g., GCN\cite{kipf2016semi}, SGC\cite{wu2019simplifying}, GAT\cite{velivckovic2017graph}, can be regarded as a particular form of the low-pass filter \cite{nt2019revisiting,xu2020graph}, which can extract the low-frequency information in the graph data and make the characterization information of a node and its surrounding context closer\cite{li2019label}. Another major stream of GRL is contrastive learning in which graph augmentations preserve the low-frequency components and perturb the middle and high-frequency components of the graph. This design hinders its application on graphs with heterophily\cite{wang2022augmentation}.

However, as shown in Fig.~\ref{fig:LocalHomophily}, there are abundant signals in the network, merely retaining the low-frequency information, which denotes the commonality in node features, is insufficient to support patterns differing from commonality. The high-frequency information, capturing the difference between node features, may be more suitable for heterophily. However, very limited GRL works have been proposed to support the embeddings for heterophily. Our previous work, PairE\cite{li2022graph} tries to encode both low-frequency and high-frequency signals among connected nodes by employing node pairs as basic embedding units. However, systematic encoding of different signals in GRL remains under-explored.


One major limitation faced by most GRL solutions lies in that one pretext task may only retain information from one perspective and can not cover the rich patterns existing in graphs\cite{jin2021automated}. To support this conclusion, we assess the quality of the embeddings generated from two different pretext tasks: the \textit{ego-task} that keeps the difference between node attributes and the \textit{agg-task} that keeps the commonality signals between a node and its surrounding neighbors, which explore high-frequency and low-frequency signals respectively. Fig.~\ref{fig:High} and \ref{fig:low} clearly illustrate an important fact: embeddings from the two tasks have the exact opposite performance in Pubmed and Wisconsin. Specifically, Fig.~\ref{fig:low} shows that when a network exhibits heterophily(Wisconsin), embeddings from high-frequency signals (ego-task) perform much better than low-frequency signals (agg-task), while under homophily settings(Cora) are opposite, as shown in Fig.~\ref{fig:High}. As already discussed, real-world graphs usually display a complex and typically mixture of patterns. Thus,  it is essential to extract different frequency signals simultaneously.

Motivated by the above discussed limitations with one pretext task, this paper proposes a novel Multi-view Graph Encoder(MVGE) framework that can effectively condense different types of signals into the embeddings with multiple novel pretext tasks. In order to tackle the problems of the semantic difference among embeddings learned from different tasks, we also propose a novel design to transform those embeddings into one uniform semantic space. Our contribution is summarized as follows:

\begin{itemize}
    \item GRL beyond the homophily assumption. We analyze the efficacy and limitations of different pretext tasks in terms of their capabilities in capturing different types of signals existing in graphs. We show that different pretext tasks can be only effective for graphs with either homophily or heterophily and innovatively point out the necessity of learning from multiple perspectives. 
    
    \item Capturing different types of information in the graph. This paper designs two pretext tasks that capture the graph's high and low-frequency signals within node features. More specifically, the paper proposes two pretext tasks: the ego-task learns the distributions of node features with rich high-frequency signals, and the agg-task learns the distributions of the smoothed aggregated neighbor attributes with low-frequency signals. Two separated autoencoders are proposed to support those two tasks.
    \item Effective integration of embeddings from multiple perspectives. The direct use of multi-task learning might not improve final embedding quality due to their diverse semantic meanings. This paper designs and implements MVGE that can merge different embeddings with an adjacent reconstruction task which can translate embeddings in different semantic spaces into one coherent space.
    
\end{itemize}





In order to evaluate the quality of generated embeddings, we compare MVGE with nine state-of-the-art baselines on eight real-world networks covering the full spectrum of low-to-high global homophily. Extensive comparative experiments, including node classification, link prediction, and pairwise node classification, validate that MVGE has advantages over state-of-the-arts and gains significant performance improvement in heterophily while being competitive in homophily.

\section{Related Work}

  In this section, we introduce the graph representation learning under the homophily or heterophily assumptions. 

\subsection{GRL under Homophily Assumption.}

\noindent\textbf{Matrix Factorization \& DeepWalk.} Early unsupervised methods for learning node representation are traditionally based on matrix factorization and random walks. Matrix factorization methods calculate losses with handcrafted similarity metrics to build vector representations for each node with latent features\cite{yang2015network,zhang2019prone}. The inspiration for random walk-based unsupervised methods for learning node representation comes from the effectiveness of the NLP method. DeepWalk\cite{perozzi2014deepwalk} and Node2Vec\cite{grover2016node2vec} optimize node embeddings by matching nodes' co-occurrence rate on short random walks over graphs. They are shown to over-emphasize proximity information at the expense of structural information\cite{ribeiro2017struc2vec,velickovic2019deep}. Also, they are limited to preserving the similarity of adjacent nodes and cannot extend to heterophily settings. 

\noindent\textbf{AutoEncoder-based.} Graph autoencoders, e.g., GAE and VGAE\cite{kipf2016variational} and their follow-up work ARVGE\cite{pan2018adversarially} with an adversarial regularization framework use two-layer GCN as their encoder and consider that impose the topological closeness of nodes in the graph structure on the latent space by predicting the first-order neighbors. GAEs over-emphasize proximity information and are also based on the assumption that connected nodes should be more similar. 

\noindent\textbf{Unsupervised GNNs.} Unsupervised GNN-based methods\cite{hamilton2017inductive,you2019position}, on the other hand, propagate features and aggregate them within various graph neighborhoods via different mechanisms (e.g., averaging, LSTM), where the node representations evolve over multiple rounds of propagation with becoming prohibitively similar. Uniform aggregation and update in GNNs ignore the difference in information between similar and dissimilar neighbors. On heterophilic graphs, discriminative node representation learning yearns for distinguishable information with diverse message passing\cite{zhu2020beyond}\cite{zheng2022graph}.

\noindent\textbf{Contrastive methods.} Contrastive methods aim to learn representations by maximizing feature consistency under different augmentation views\cite{herzig2020learning}. For example, DGI\cite{velickovic2019deep} employs the idea of Deep InfoMax\cite{hjelm2018learning} and considers both local and global information during discrimination. MVGRL\cite{hassani2020contrastive} proposed to train graph encoder by maximizing MI between representations encoded from first-order neighbors and a graph diffusion. MERIT\cite{jin2021multi} utilized Siamese networks\cite{chen2021exploring} and proposed cross-views and cross-network contrastive objectives to maximize the similarity between node representations from local and global perspectives. These methods can, to some extent, mitigate the impacts of heterophily. However, most of those solutions only augment the low-frequency signals between nodes and perturb the middle and high-frequency components of the graph. It contributes to the success of GCL algorithms in homophily but hinders its application in heterophily\cite{wang2022augmentation}. 


\subsection{Learning beyond Homophily}

Due to the lack of node labels, graph representation learning beyond homophily remains largely unexplored. Existing works are mainly limited to the semi-supervised GNNs for heterophily. The common philosophy behind these works is weakening the smoothing effect. For example, MixHop\cite{abu2019mixhop} and H2GCN\cite{zhu2020beyond} combine multi-scale information by concatenation to modify the smoothed node representations based on local information, while FAGCN\cite{bo2021beyond} and GPRGNN\cite{chien2020adaptive} relax the edges with positive values, which tends to induce the smoothing effect, to possess real values (which can be either positive or negative). 
GNN-LF/HF~\cite{zhu2021interpreting} designs filter weights from the perspective of graph optimization functions, which can simulate high- and low-pass filters and LINKX~\cite{lim2021large} separately embeds node features and graph topology. After that, the two embeddings are combined with MLPs to generate node embeddings.
However, these methods are semi-supervised and rely on the node labels to define the homophily level and guide the learning processes. Actually, different attributes possess diverse characteristics in unsupervised settings. 

GRL adopting an unsupervised/self-supervised paradigm still remains challenging due to the uncertainty in defining a learning objective.  Selene~\cite{zhong2022unsupervised} proposes a dual-channel feature embedding mechanism to fuse node attributes and network structure information and leverage a sampling and anonymization strategy to break the implicit homophily assumption of existing embedding mechanisms. However, Selene needs to extract the high-order subnetwork ($r$-Ego) corresponding to each node, which will significantly increase the computational and space burden. Our previous work, PairE\cite{li2022graph} preserves both information for homophily and heterophily by going beyond the localized node view and utilizing paired node entities as the basic embedding unit. It can, to some extent, maintain high-frequency signals. But it demands hand-craft design to translate pair-based embeddings to node embeddings for typical downstream tasks. AF-GCL\cite{wang2022augmentation} leverages the features aggregated by GNN to construct the self-supervision signal, which prevents perturbing middle and high-frequency information due to the heavily relying on graph augmentation and is less sensitive to the homophily degree. However, AF-GCL still aims to learn similarities between nodes rather than keep high-frequency signals in node features. In the settings of unsupervised GRLs, how to effectively encode different frequency signals into embeddings and make the graph representation can effectively support a wealth of unknown downstream tasks remains largely unexplored.

\section{Notation and Preliminaries}

Let $\mathcal{G} = (\mathcal{V};\mathcal{E})$ be an undirected, unweighted graph with node set $\mathcal{V}$ and edge set $\mathcal{E}$. For each node $v$, there contains a unique class label $y_v$. We denote a general random walk rooted at vertex $v$ as $W(v)$ and the $i$ hops/steps random walk as $W_i(v)$. For example, $W_1(v) = {u : (u,v) \in \mathcal{E}}$ is one of the immediate neighbors of $v$. $N(v)$ denotes the 1-hop neighbors of node $v$. We represent the graph by its adjacency matrix $A \in {0,1}^{N \times N}$ and its node feature matrix $X \in \mathcal{R}^{N \times F}$, where the vector $x_v$ corresponds to the ego-feature of node $v$ and ${x_u : u \in W(v)}$ to features of the node in its random walk path.

\subsection{Global and Local Homophily}
Here, we define the metrics to describe the essential characteristics of a graph. There exist multiple metrics, e.g., the label smoothness defined in~\cite{hou2019measuring} or the network assortativity in~\cite{newman2003mixing}. Here, we adopt the edge homophily ratio $h^{global}$,  the fraction of edges which connect nodes that have the same class label(i.e., intra-class edges), which gives an overall trend for all the edges in the graph. This metric is firstly defined in H2GCN\cite{zhu2020beyond}. 

\noindent\textbf{Definition 1}\textit{ (Global Homophily).} The edge homophily ratio $h^{global}$

\begin{equation}
h^{global} = \frac{|\{(u,v):(u,v)\in \mathcal{E} \wedge y_u = y_v\}|}{|\mathcal{E}|}
\label{equ:local}
\end{equation}

When the global edge homophily ratio $h^{global} \to 1$, it shows that the graph has strong homophily. In contrast, graph with strong heterophily(i.e., low/weak homophily) have small global edge homophily ratio $h^{global} \to 0$.

However, as discussed in the introduction, this metric only represents the global trend of a graph. In different parts of real-world graphs, different homophily degrees can be observed. Here, we explicitly introduce a new metric that measures the homophily degree from a node's local view.

\noindent\textbf{Definition 2} \textit{(Local Homophily).} The local homophily is the fraction of nodes in node $v$'s immediate neighbors that have the same class label as node $v$:

\begin{equation}
    h^{local}_v = \frac{|\{(u,v):(u,v) \in \mathcal{E} \wedge y_u = y_v\}|}{|\{(u,v):(u,v) \in \mathcal{E}\}|}
\end{equation}

Rather than $h^{global}$ that represent global trend with one scalar, the local edge homophily ratio $h^{local}_v$ is node-specific. When $h^{local}_v \to 1$, it indicates that node $v$ connects to surrounding nodes with the same label, while node $v$ with strong heterophily $h^{local}_v \to 0$.

\subsection{Node Features in Ego and Aggregation Views}
As pointed out by previous works\cite{zhu2020beyond,bo2021beyond}, in graphs with strong homophily, the attribute aggregation operation in GNNs can usually intensify the low-frequency signals. While in graphs with strong heterophily, the node attributes themselves and MLP can often achieve SOTA performance as those attributes contain a distinctive node pattern for classification. As we do not have any labels and downstream tasks might rely on different types of signals, we go beyond the traditional homophily assumption and innovatively propose two views to capture the low- and high-frequency information existing in the node features: specifically, ego features and aggregated features. 

\noindent\textbf{Definition 3} \textit{(Node-specific Ego features).} We use the nodes' original features as ego features: 
\begin{equation}
X^{ego} = \{x_v^{ego}:x_v^{ego}=x_v,v \in \mathcal{V} \}
\end{equation}
It contains the unique information of the node itself, which can preserve the information difference between nodes to the greatest extent, which is the high-frequency signals.

\noindent\textbf{Definition 4} \textit{(Walk-based Aggregated features).}
Merely using the ego features, the noise in them will cause the learned representations to deviate from reality. To reduce the impacts of the noise, similar to the GNN-based solutions, we define an aggregation function to aggregate the surrounding context, denoted as $X^{agg}$. One important requirement is the aggregated feature should be able to intensify the low-frequency signals and/or the mean-smoothed high-frequency signals around a node $v$. Furthermore, to some extent, the aggregated features should represent the local structure of node $v$. Here, the aggregated neighbor feature for each node $v$ is given as: 

\begin{equation}
\label{eq:walk_agg}
x^{agg}_v = AGGR(average(x_{u_{i}}:u_i\in W_i(v)),average(x_{u_{j}}:u_j\in W_j(v)),...)
\end{equation} 

\begin{wrapfigure}{l}{0.35\columnwidth}
  \centering
  \includegraphics[width=0.34\textwidth]{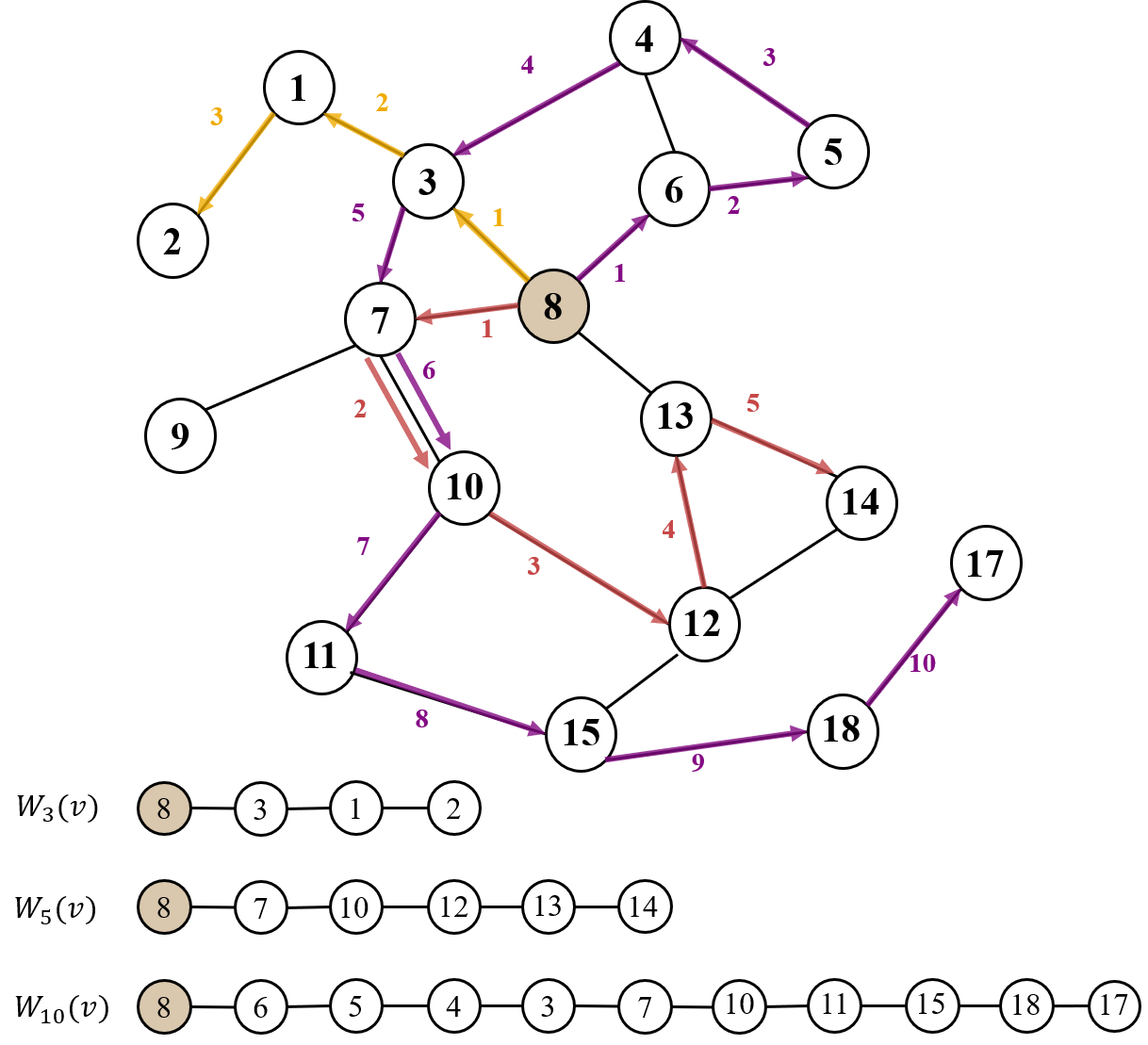}
  \caption{DeepWalk-based feature aggregation}
  \label{fig:deepwalk}

\end{wrapfigure}

Specifically, for each node $v$, we generate several paths through unbiased random walks with unequal steps, e.g., $W_i(v)$,$W_j(v)$. Here, i and j represent the length of the path. The aggregated features from each path are calculated with the features-wise average for all the nodes along the path. Figure~\ref{fig:deepwalk} shows the three paths starting from node 8 with lengths 3, 5, and 10. Then, we mixed the aggregated features of each random walk path by concatenation. This design is based on the following observation. (1) multiple random walks help to capture mixing latent information from neighbors at various distances. (2) As verified in H2GCN, in heterophily settings, although the labels of immediate neighbors tend to be different from the ego node, the higher-order neighborhoods may have the same class of the ego node and can provide more relevant context. Thus, by higher-order random walk and aggregating information beyond the immediate neighbors, the similarity between nodes with structural and semantic similarity can be preserved.

\subsection{Feature-wise Low and High-frequency Signals}

Here we hope to explain low-frequency and high-frequency signals and their relationship to ego and aggregated features. 

\begin{wrapfigure}{r}{0.35\columnwidth}
  \centering
  \includegraphics[width=0.28\textwidth]{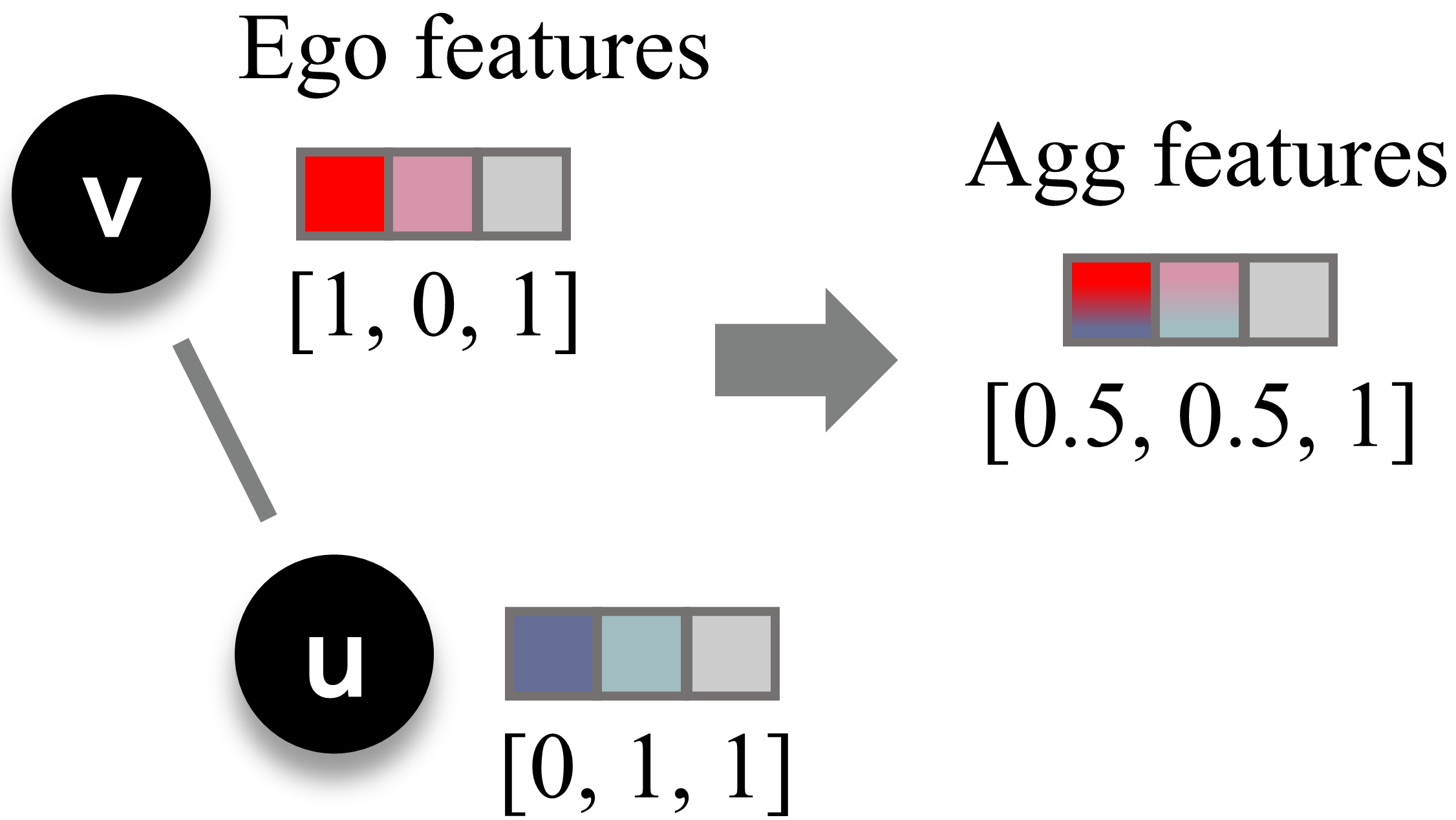}
  \caption{Different colors denote different dimensions of features. Different values represent high-frequency signals in this dimension, otherwise low-frequency signals.}
  \label{fig:frequency}
\end{wrapfigure}

\textbf{Definition 5} \textit{(Low/High-frequency signals)} According to the definition in \cite{bo2021beyond, zhu2021interpreting}, the low-frequency signal represents the similarity information between node features, and the high-frequency signal represents the difference information between node features. The greater the similarity of node feature information, the stronger the low-frequency signal transmitted, otherwise, the high-frequency signal dominates.

Actually, different features possess diverse characteristics and can not be directly compared. To explicitly model the diversity, we define low and high-frequency signals from a single feature dimension. As shown in Fig.~\ref{fig:frequency}, taking a central node $v$ and any node $u$ as an example, the 1st/2nd-dim features with significant differences can be seen as existing high-frequency signals, while the 3rd-dim features exist as low-frequency signals. The agg features, which are the aggregated features from neighbor node $u$ and central node $v$, smooth high-frequency signals and identify low-frequency signals.

\section{Multi-view Graph Encoder}
MVGE is inspired by the approaches that explore the combination of multiple pretext tasks to learn different signals\cite{jin2021multi,hassani2020contrastive,li2022graph,jin2021automated}. This framework includes two novel pretext tasks supported with multi-view parallel AutoEncoders(Section~\ref{sec:SeperatedEncoders}), a novel solution to integrate embeddings from two views into one coherent semantic space(Section~\ref{sec:integration}).

\begin{figure}[tbp]
  \centering
  \includegraphics[width=0.9\textwidth]{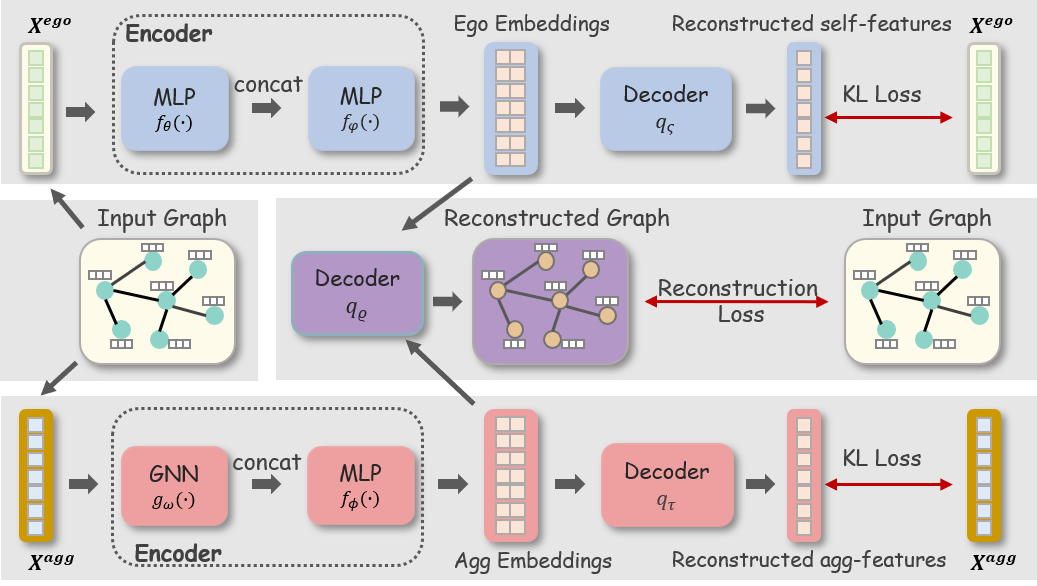}
  \caption{Overall framework of MVGE. MVGE tasks two augmentation views as input, i.e., ego features  $X^{self}$ with high-frequency signals and aggregated features $X^{agg}$ with low-frequency signals. From these two views, we design two feature reconstruction pretext tasks with KL loss. By employing two separated encoders, MVGE generate embeddings including different semantic information: (1) linear encoder for $X^{self}$, with two single layer MLP $f_{\theta}(\cdot)$ and $f_{\varphi}(\cdot)$; and (2) GNN encoder for $X^{agg}$, with a two-layer GCN $g_{\omega}(\cdot)$ and a single layer MLP $f_{\phi}(\cdot)$; (3) Similar to the ”add skip connections” operation, we 'concat' the inputs and the intermediate representation at the final layer. With the above preparations, MVGE starts the training process controlled by three terms loss function. $q_{\varsigma}$ and $q_{\tau}$ denotes a linear layer-based decoder. $q_{\varrho}$ is an inner product decoder.}
  \Description{overall framework of MVGE}
  \label{fig:framework}
\end{figure}


\subsection{Separated Encoders for Different Views}
\label{sec:SeperatedEncoders}
Traditional self-supervised learning feeds different data augmentations into a single embedding space or two embedding spaces with shared parameters, which interferes with learning different features. Motivated by this limitation, our model learns to construct separated embedding sub-spaces for different views, which avoids mixing low frequency and high-frequency signals and allows for more expressiveness for different frequency signals. Furthermore, we devise two similar pretext tasks to learn signals from the two views for each encoder. 

\subsubsection{Linear Encoder for Ego-features.}
To preserve high-frequency information in ego features, we propose to replace the generic GCN encoder with a linear encoding scheme. As illustrated in Figure~\ref{fig:framework} in blue block, the ego features are encoded by two single layer MLP $f_{\theta}(\cdot)$ and $f_{\varphi}(\cdot)$. Different from the standard GNN-based encoder, a single linear transformation of the features does not smooth the feature signal of the surrounding neighbors during propagation, thus maximizing the preservation of the high-frequency information of the nodes themselves and keeping node representations from becoming indistinguishable. Besides, similar to the "add skip connection" operation introduced in ResNet\cite{he2016deep}, which is later adopted in jumping knowledge networks for graph representation learning\cite{xu2018representation}, we (2) concatenate the original inputs and the intermediate representations at the final layer, to preserve the integrity and personalization of the input information as much as possible.


\subsubsection{GNN Encoder for Agg-features.}

For the aggregated features from neighbors, we want to learn the commonality among neighbors. Thus, we adopt a GCN encoder for the aggregated features. In general, GCN updates node representations by aggregating information from neighbors, which can be seen as a special form of low-pass filter\cite{wu2019simplifying,li2019label} and can retain the commonality in the neighbors. As shown in Figure \ref{fig:framework} in red block, the agg features are encoded by a two-layer GCN $g_{\omega}(\cdot)$ and a single layer MLP $f_{\phi}(\cdot)$.


\subsubsection{Learning in Different Views}
\label{sec:views}
Generally, the self-supervision task are normally use the reconstruction loss\cite{velivckovic2017graph} or the contrast loss\cite{velickovic2019deep}. To capture both the low-frequency and high-frequency signals in the node attribute, we utilize Kullback-Leibler divergence~\cite{kullback1997information} and propose two feature reconstruction pretext tasks to learn signals from two views.

\noindent\textbf{Ego Feature Reconstruction.} To capture low-frequency signals in the embedding, MVGE adopts the ego feature reconstruction pretext task from the ego feature view. This task is designed to recover the ego features $X^{ego}$. As each node usually has its own specific feature vector, it usually has distinctive information that differs itself from other nodes. As pointed out by many previous GNNs\cite{zhu2020beyond,bo2021beyond,li2022graph}, the integration of ego-features can help the embeddings keep high-frequency signals and prevent the node representation from tending to be similar during the model training process.

Specifically, we use $Z^{ego} = q_{\varsigma}(H)$ to denote the output embedding of decoder $q_{\varsigma}$. Aiming to pull closer to the representations from the original features, we use the ego feature reconstruction loss with the Kullback-Leibler divergence $\mathcal{L}_{ego}$. $\mathcal{L}_{ego}$ calculate the distribution difference between $P^{ego}$ and $Q^{ego}$, where Q is the distribution of the reconstructed features $Z^{ego}$ and $P^{ego}$ is the original distribution of $X^{ego}$:

\begin{equation}
\label{L_ego}
    \mathcal{L}_{ego} = KL[P^{ego}||Q^{ego}] = \sum_k\sum_j p_{kj}^{ego} log{\frac{p_{kj}^{ego}}{q_{kj}^{ego}}}
\end{equation}

\noindent where $k$ represents the $k$-th node in $\mathcal{V}$, $j$ represents the $j$-th feature. $q_{kj}^{ego}$ and $p_{kj}^{ego}$ are the distribution possibility and the target distribution of the $j$-th features of the $k$-th node, respectively:
\begin{equation}
    q_{kj}^{ego} = \frac{exp(Z_{kj}^{ego})}{\sum_j exp(Z_{kj}^{ego})}, p_{kj}^{ego} = \frac{exp(X_{kj}^{ego})}{\sum_j exp(X_{kj}^{ego})}
\end{equation}

\noindent\textbf{Aggregated Feature Reconstruction.} 
Similar to the ego feature reconstruction task, this task reconstructs the aggregated features $X^{agg}$ and uses a loss function based on KL divergence:
\begin{equation}
\label{L_agg}
    \mathcal{L}_{agg} = KL[P^{agg}||Q^{agg}] = \sum_k\sum_j p_{kj}^{agg} log{\frac{p_{kj}^{agg}}{q_{kj}^{agg}}}
\end{equation}

\noindent where $q_{kj}^{agg}$ and $p_{kj}^{agg}$ can be calculated as the same way like $q_{kj}^{ego}$ and $p_{kj}^{ego}$.

By reconstructing the smoothed node neighborhood features, the model can capture low-frequency information,  which helps avoid node representations deviating from reality in multiple rounds of learning guided by self-feature reconstruction tasks.






\begin{wrapfigure}{r}{0.50\columnwidth}
  \centering
  \begin{subfigure}{0.24\textwidth}
  \centering
  \includegraphics[width=\textwidth]{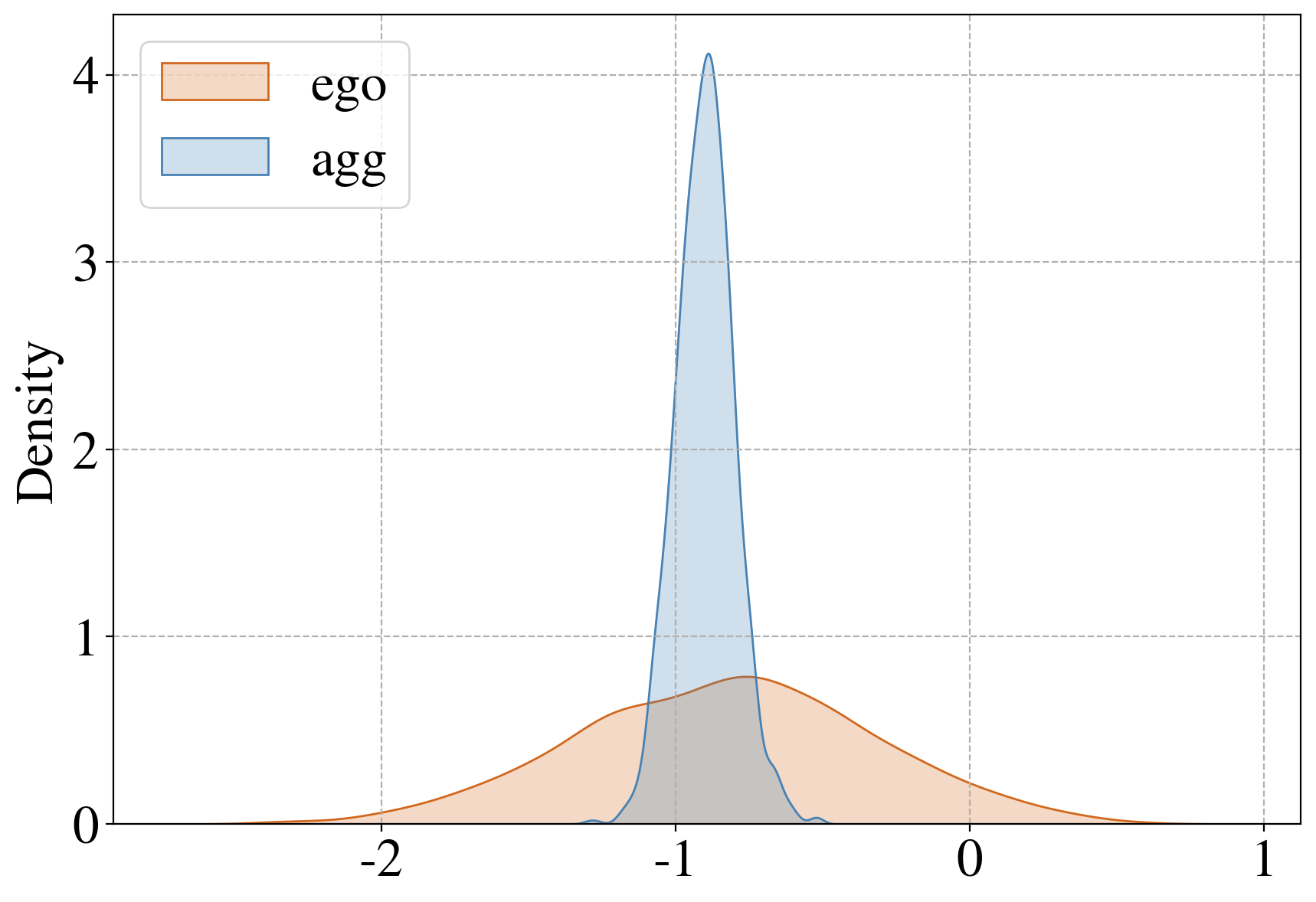}
  \caption{$h^{global}=0.1$}
  \label{fig:Embedding1}
 \end{subfigure}
 \begin{subfigure}{0.24\textwidth}
  \centering
  \includegraphics[width=\textwidth]{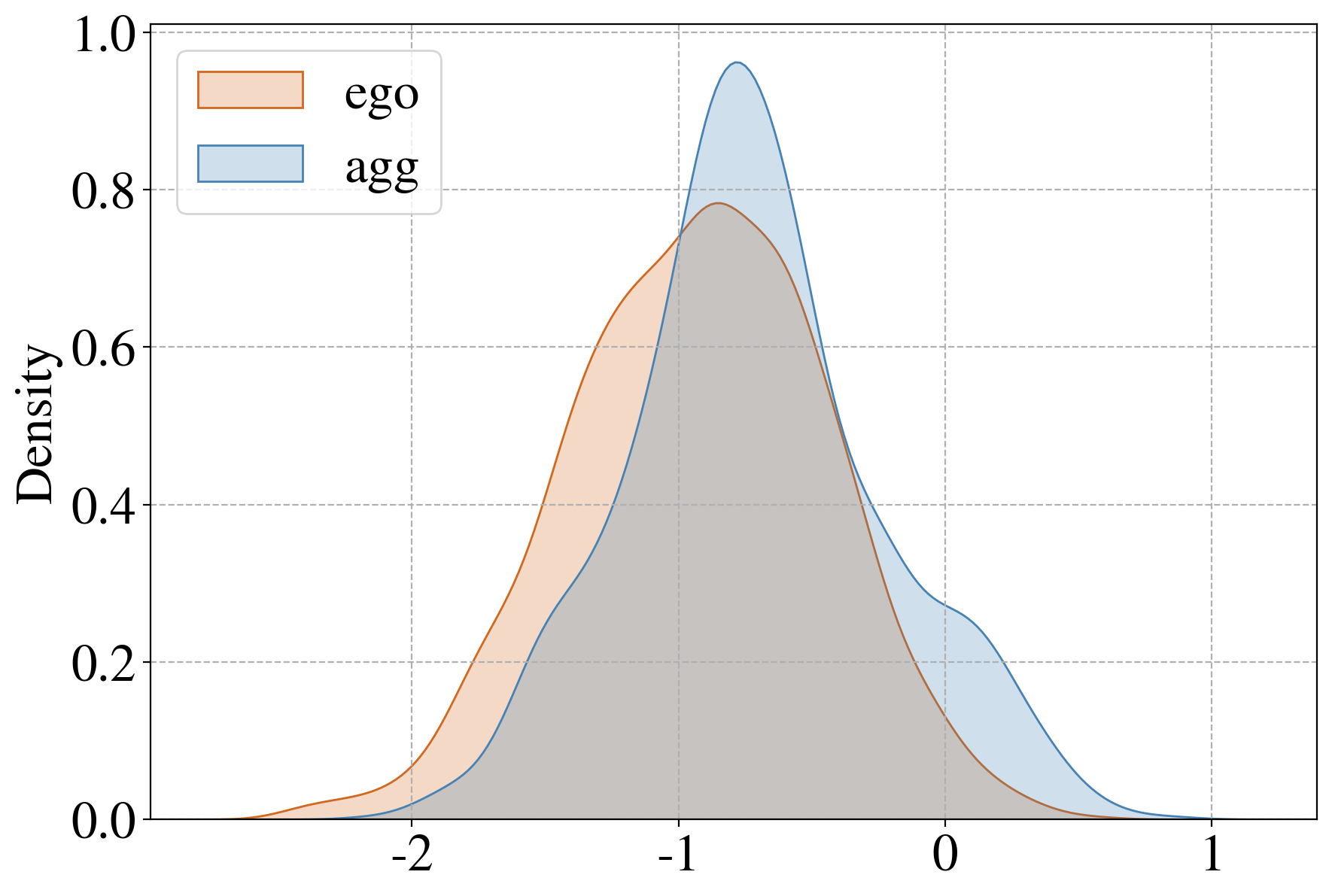}
  \caption{$h^{global}=0.9$}
  \label{fig:Embedding3}
 \end{subfigure}
 \caption{Embedding distribution of syn-cora dataset under different global homophily in the first dimension. Others have similar distributions}
 \label{fig:Embedding}
\end{wrapfigure}

\subsection{Learning different signals.} Fig.~\ref{fig:Embedding} shows the embedding distributions in the first dimension generated by respective feature reconstruction tasks on the syn-cora with low (0.1) and high (0.9) homophily ratios. As we can see from this figure, as the NN optimizer algorithms normally generate weights according to Gaussian distribution, all those embeddings basically follow a Gaussian distribution, however, with different standard deviations $\sigma$. Here, we can observe two important facts. 1) Under different global homophily, $\sigma_{ego}$ is bigger than $\sigma_{agg}$. This result demonstrates that the ego feature reconstruction task preserves more high-frequency signals. In comparison, in the aggregated feature reconstruction task, due to its smoothing operations in both deepwalk generation and GCN, the embeddings generated by the agg-task mainly contain low-frequency signals with small $\sigma$. (2) Embeddings from the two tasks become similar in strong homophily settings while dissimilar in heterophily settings. In datasets with low homophily, nodes with different labels tend to be connected, and the nodes are very different from their surrounding neighbors. Thus, the ego embeddings are quite different from agg embeddings.
In datasets with strong homophily($h^{global}=0.9$), the similarity between linked nodes is very high, and the feature difference between the connected node(high-frequency signals) is relatively less. Thus, as shown in Fig.~\ref{fig:Embedding3}, these two tasks generate almost identical distributions. Fig.~\ref{fig:Embedding1} and ~\ref{fig:Embedding3} clearly show that the two tasks can capture different types of signals.


\subsection{Integration of multiple semantic embeddings}
\label{sec:integration}

Our model learns to capture low-frequency and high-frequency signals for node representations by constructing separate embedding spaces and introducing two pretext tasks. Although, as pointed in ~\cite{zhang2021survey}, learning with multiple tasks can generally improve performance compared to learning from a single task in CV, NLP, etc. However, simply using multiple pretext tasks in graph learning normally leads to unstable and poor performance in graphs\cite{standley2020tasks}. The key to this factor, we guess, is due to the fact that the learned embeddings from different pretext tasks are actually from different semantic spaces. It is hard for the ML classifier to use embeddings from different semantic spaces effectively. 



In MVGE, although the ego-task and agg-task are similar as they are both based on the feature distribution reconstruction, their semantic meanings are still quite different. To bridge the gap between two semantic spaces of different pretext tasks, we propose one additional pretext task, the adjacency matrix reconstruction task, to seamlessly integrate the embeddings from two views. Another reason to introduce this pretext task is that we need to keep a certain graph structure while the two views are generally from the feature view and lack topology information. 

\noindent\textbf{Adjacency Matrix Reconstruction.} Specifically, as shown in Fig.~\ref{fig:framework} in purple blocks, we use a traditional inner product decoder $q_{\varrho}$ to reconstruct the graph structure for link prediction, which is defined as:

\begin{equation}
\hat{A} = \sigma (HH^T)
\end{equation}

\noindent where $H$ is obtained by concatenating 'Ego Embeddings' and 'Agg Embeddings', $\hat{A}$ denotes the reconstructed adjacency matrix, and $\sigma (\cdot)$ is a non-linear activation function - we use the sigmoid activation function in MVGE. This function determines the likelihood of pairwise nodes connected in $\hat{A}$. During the training phase, our goal is to iteratively minimize a reconstruction loss by capturing the similarity between $A$ and $\hat{A}$. Thus, the objective function $\mathcal{L}_{s}$ can be formulated as a cross-entropy loss:

\begin{equation}
    \mathcal{L}_s = -\frac{1}{N^2}\sum_{i=1}^N \sum_{j=1}^N A_{ij}log\hat{A}_{ij}+(1-A_{ij})log(1-\hat{A}_{ij})
\end{equation}



\noindent\textbf{Learning with three pretext tasks.} To train our model end-to-end and learn node representations for downstream tasks, we jointly leverage both the ego and aggregated feature restoration loss and the adjacency matrix reconstruction loss. Thus, the overall object function is defined as:
\begin{equation}
    \mathcal{L} = \beta (\alpha \mathcal{L}_{ego} + (1-\alpha)\mathcal{L}_{agg}) + (1-\beta)\mathcal{L}_s
\end{equation}
\noindent where we aim to minimize $\mathcal{L}$ during the optimization, and $\alpha$ represents the weights that balance the  $\mathcal{L}_{ego}$ and $\mathcal{L}_{ego}$, $\beta$ is a balance factor that balances the averaged feature reconstruction loss and the adjacency matrix reconstruction loss.



\section{Experiments}
We choose three graph analysis tasks, i.e., node classification, link prediction, and pairwise node classification, to evaluate the effectiveness of representations learned from graphs with different global homophily. And we demonstrate the superiority of MVGE by comparison with 9 state-of-the-art techniques. Furthermore, we show the significance of designs in MVGE through rich ablation studies. The experiments are performed based on 8 popular benchmark datasets with global homophily ratio $h^{global}$ ranging from strong heterophily to strong homophily and two synthetic datasets where we can control the homophily/heterophily level.

\subsection{Experiment Settings:}

\noindent\textbf{Datasets.} 
Statistic details of those datasets are listed in Table~\ref{tab:datasets}.

\begin{itemize}
    \item Cora, Citeseer, Pubmed, Cora$\_$full\cite{namata2012query}: standard citation networks where nodes represent documents, edges are citation links, and features are the bag-of-words representation of the document.
    \item Chameleon\cite{rozemberczki2021multi}: page-page networks in Wikipedia, where nodes represent articles from the English Wikipedia, edges reflect mutual links between them. Node features indicate the presence of particular nouns in the articles and the average monthly traffic.
    \item Cornell, Texas, Wisconsin\cite{pei2020geom}: school department Web Page networks, where nodes represent web pages, and edges are hyperlinks between them. Node features are the bag-of-words representation of web pages. The web pages are manually classified into the five categories, student, project, course, staff, and faculty.
    \item Synthetic datasets. syn-cora and syn-products are two synthetic datasets where we can control the homophily/heterophily level\cite{zhu2020beyond}. Node features are generated by sampling nodes with the same class label from real-world datasets Cora and ogb-products.
\end{itemize}

\begin{table}[!htb]
  \begin{center}
  \caption{The statistics of the datasets.}
  \label{tab:datasets}
    \begin{tabular}{lccccc}
    \hline
    \textbf{Dataset} & \textbf{Nodes} & \textbf{Edges} & \textbf{Features} & \textbf{Classes} & $h^{global}$\\ 
    \hline
    Cora         &2,708 &5,278 &1,433 &7  &0.81  \\
    Citeseer     & 3,327  &4,676 &3,703 &6  &0.74 \\
    Pubmed       &19,717 &44,327 &500 &3 &0.8\\
    Cora$\_$full &19.793 &63,421 &8710 &70 &0.57\\
    Chameleon    &2,277  &31,371  &2325 &5 &0.23\\
    Cornell &183  &277 &1,703 &5 &0.3 \\
    Texas &183 &279 &1,703 &5 &0.11  \\
    Wisconsin &251 &450 &1,703 &5 &0.21\\
    syn-cora &1,490 &2,968 &1,433 &5 &[0.1,0.3,0.5,0.7,0.9]\\
    syn-products  &10,000 &59,648 &101 &10 &[0.1,0.3,0.5,0.7,0.9]\\
    \hline
    \end{tabular}
    \end{center}
\end{table}

\noindent\textbf{Baselines.} 
\label{sec:baselines}
In comparative experiments, we compare MVGE with 9 self-supervised models with open source code, including two contrastive models learning from multi-views with several training objectives, MERIT and MVGRL, and one beyond homophily assumption, PairE. We aim to provide a rigorous and fair comparison between different models on each dataset by using the same dataset splits and training procedure. The introduction of the comparison method is as follows:


\begin{itemize}
\item DeepWalk\cite{perozzi2014deepwalk}:  A GRL algorithm based on random walk, utilizing Word2Vec and optimizing node embeddings by matching the co-occurrence rates of nodes on short random walk paths over graphs.
\item SAGE-U\cite{hamilton2017inductive}: An unsupervised variant of GraphSAGE generates node embeddings by sampling a fixed number of neighbors for each node and aggregating their features.
\item DGI\cite{velickovic2019deep}: A contrastive learning method via maximizing mutual information between local and global representations for embedding learning.
\item GAE\cite{kipf2016variational}: The first algorithm to process graph-structured data with autoencoder, which captures the structural information of the graph by reconstructing the adjacency matrix.
\item ARVGE\cite{pan2018adversarially}: An autoencoder-based solution that augments GAE by adding an adversarial module on the obtained embeddings to learn more robust graph representations, thereby improving the model's performance on sparse and noisy graph data.
\item P-GNN\cite{you2019position}: A new class of GNNs for computing position-aware node embeddings, which capture positions/locations of nodes with respect to the anchor nodes by computing the distance of a given target node to each anchor-set.
\item MERIT\cite{jin2021multi}: A multi-view self-supervised model that learns node representations by enhancing Siamese self-distillation with multi-scale contrastive learning. \item MVGRL\cite{hassani2020contrastive}: A multi-view self-supervised model that learns node representations and graph level representations by contrasting structural views of graphs.
\item PairE\cite{li2022graph}: A multi-task unsupervised model to preserve both information for homophily and heterophily by going beyond the localized node view and utilizing paired nodes entities with richer expression powers. 
\end{itemize}


\noindent\textbf{Experiment Settings.}
\label{sec:settings} 
For MVGE and all state-of-the-art baselines, we use the Adam optimizer with a learning rate of 0.01, and the embedding dimension of both embeddings in MVGE and node embeddings for other benchmarks is fixed to 128. All models are run 10 times, and the experiment results are the mean values with standard deviation. Other parameters are set as the default settings in the corresponding papers. For the walk-based aggregated features in MVGE, we generate 3 random walk paths for each node and set them with deep walk lengths as 3, 5, and 10 respectively. We run 200 epochs for each trainning. 


\noindent\textbf{Downstream Tasks.} The experimental settings for different downstream tasks are as follows:

\noindent\textit{Node classification.} For this task, all the nodes are used in the embedding generation process. To balance the effect of three reconstruction tasks, we perform Optuna\cite{akiba2019optuna} for efficient hyper-parameter search and tune $\alpha$ and $\beta$ between 0 and 1 with a step size of 0.1. Then, the graph encoder is trained with the obtained loss weights. To evaluate the trained graph encoder, we adopt a linear evaluation protocol by training a separate logistic regression classifier on top of the learned node representations with the one-vs-rest strategy. For all of the datasets, we use 30\% of nodes to construct the training set, while using the remaining nodes as the testing set. We repeat the experiments ten times and report the average Micro-F1-score with the standard deviation.

\noindent\textit{Link prediction.} The models are trained on an incomplete version of these datasets where parts of the citation links(edges) have been removed, while all node features are kept. Specifically, we use 15\% existing links and an equal number of nonexistent links as test sets. The remaining 85\% existing links and an equal number of randomly generated nonexistent links were used as the training set.

\noindent\textit{Pairwise node classification.} For the pairwise node classification task, we predict whether a pair of nodes belong to the same class or not. In this case, a pair of nodes that belong to the same class is a positive example and the label will be marked as 1 otherwise as 0\cite{you2019position}. Specifically, we randomly generate the same number of positive and negative samples and follow the same dataset split ratio as link prediction.

For both the link prediction task and pairwise node classification task, the paired node embeddings are calculated from the source and target node embedding with the L2 operator. We report the averaged ROC\_AUC score with the standard deviation on the test set over 10 runs with different random seeds and train/test split.

\subsection{Results on Synthetic Datasets}
Using synthetic graphs syn-cora and syn-products with various global homophily ratio $h^{global}$, we show the effectiveness of MVGE and the significance of diverse pretext tasks through node classification comparison task and ablation studies of variants of MVGE.

\begin{figure}[htbp]
\centering
\begin{subfigure}{0.32\textwidth}
  \centering
  \includegraphics[width=\textwidth]{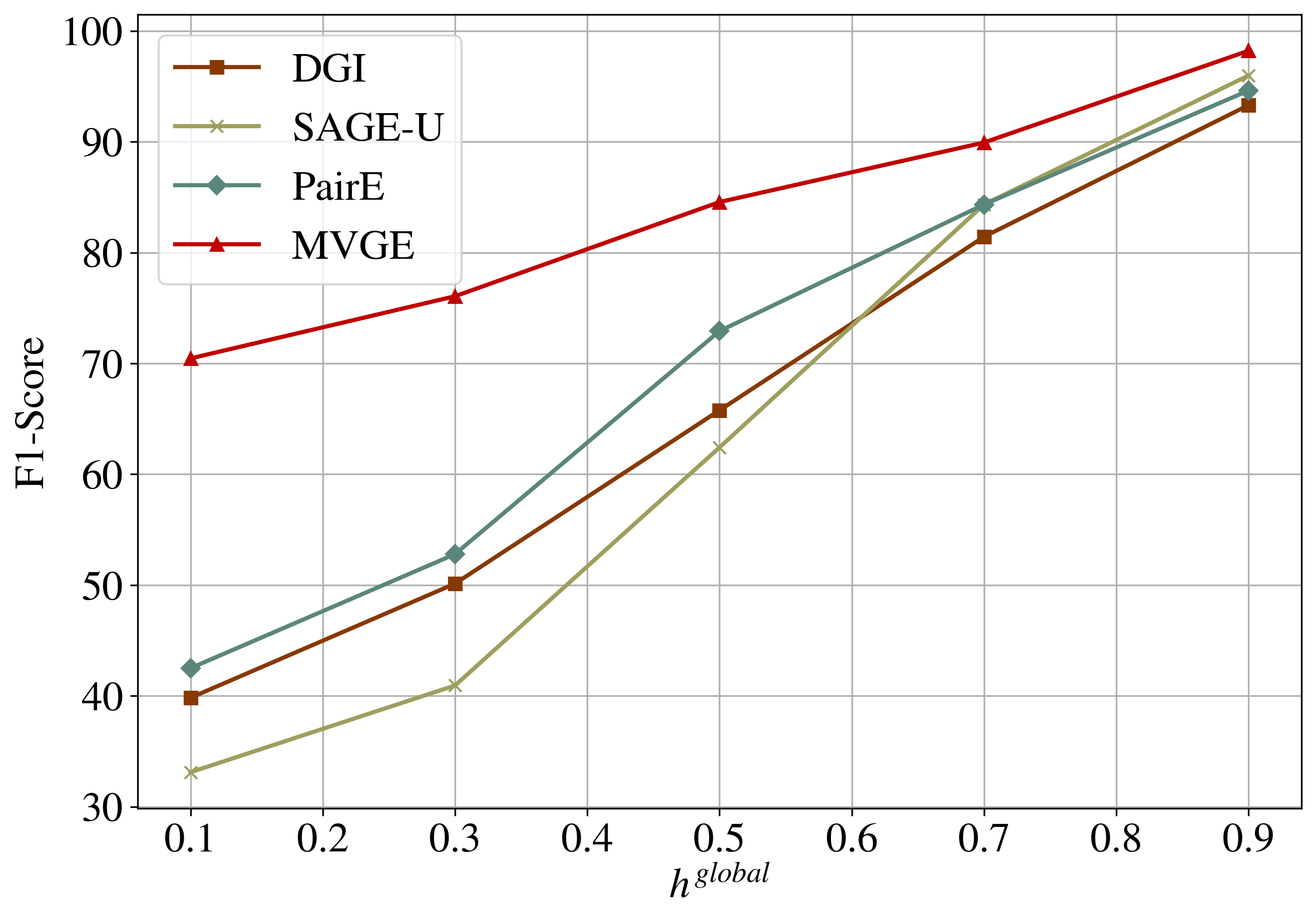}
  \caption{Cp. on syn-cora}
  \label{fig:CompOfSyn1}
 \end{subfigure}
 \begin{subfigure}{0.32\textwidth}
  \centering
  \includegraphics[width=\textwidth]{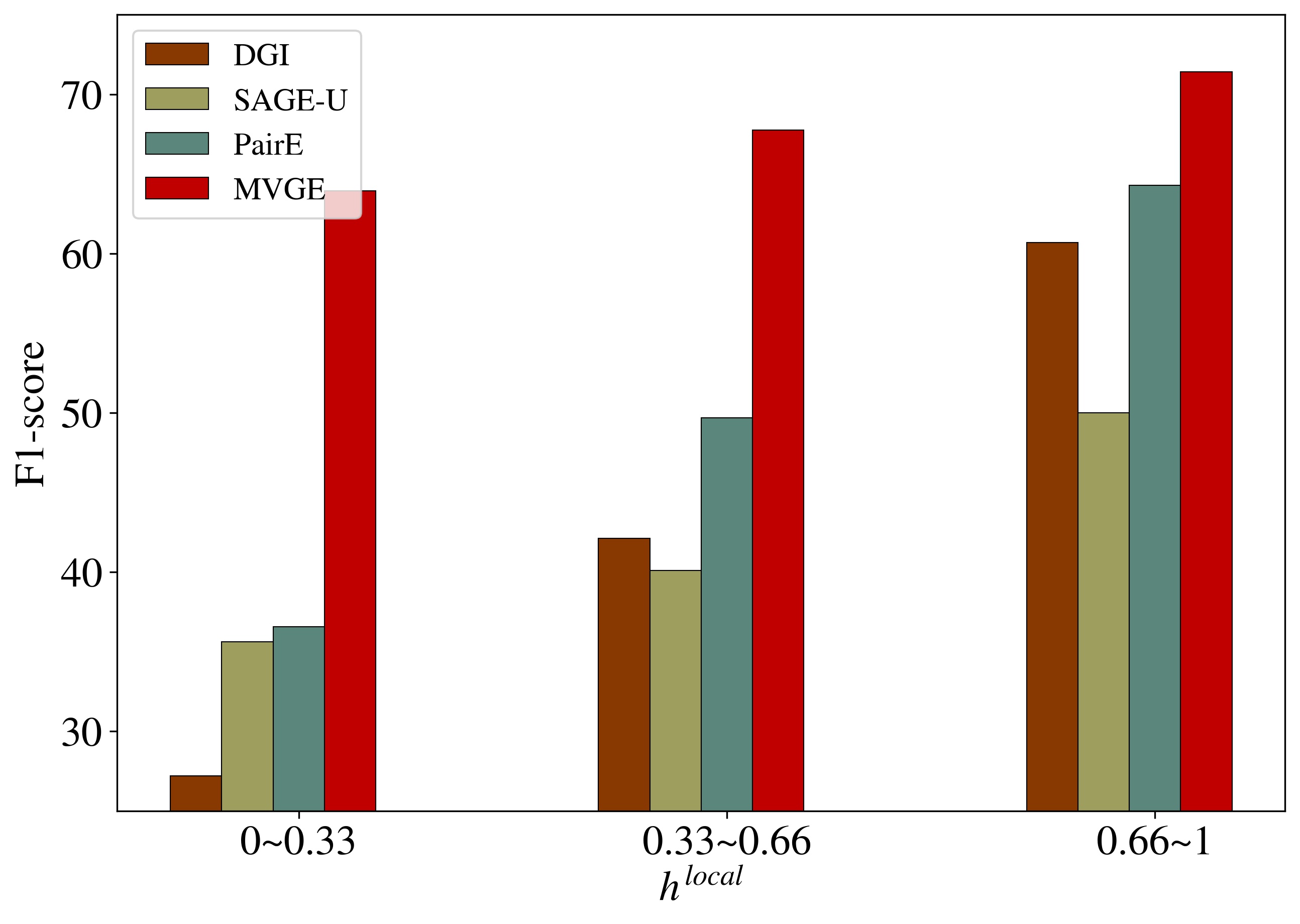}
  \caption{Cp. on syn-cora($h^{global}=0.1$)}
  \label{fig:CompOfSyn2}
 \end{subfigure}
 \begin{subfigure}{0.32\textwidth}
  \centering
  \includegraphics[width=\textwidth]{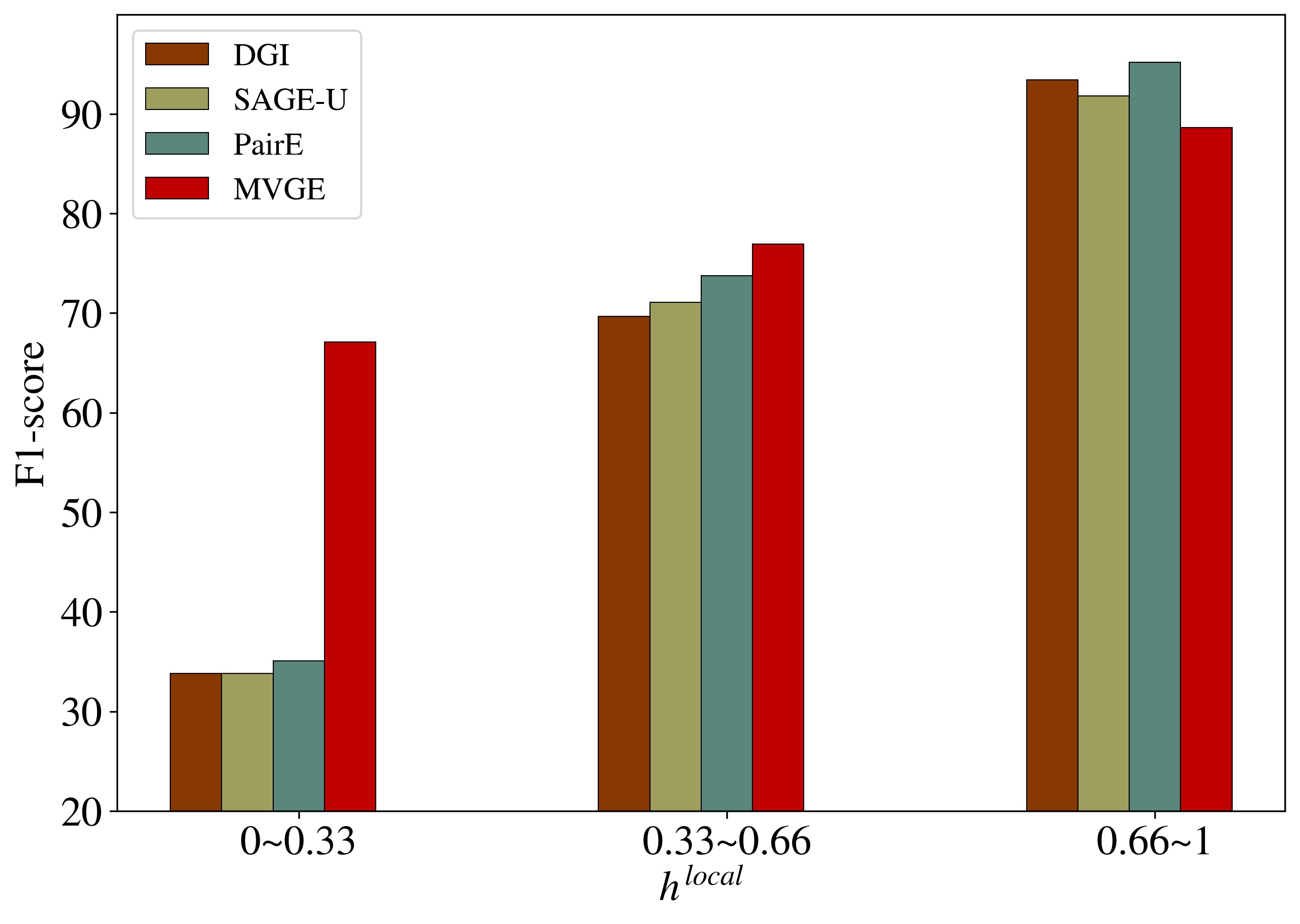}
  \caption{Cp. on syn-cora($h^{global}=0.7$)}
  \label{fig:CompOfSyn3}
 \end{subfigure}
 \\
 \begin{subfigure}{0.32\textwidth}
  \centering
  \includegraphics[width=\textwidth]{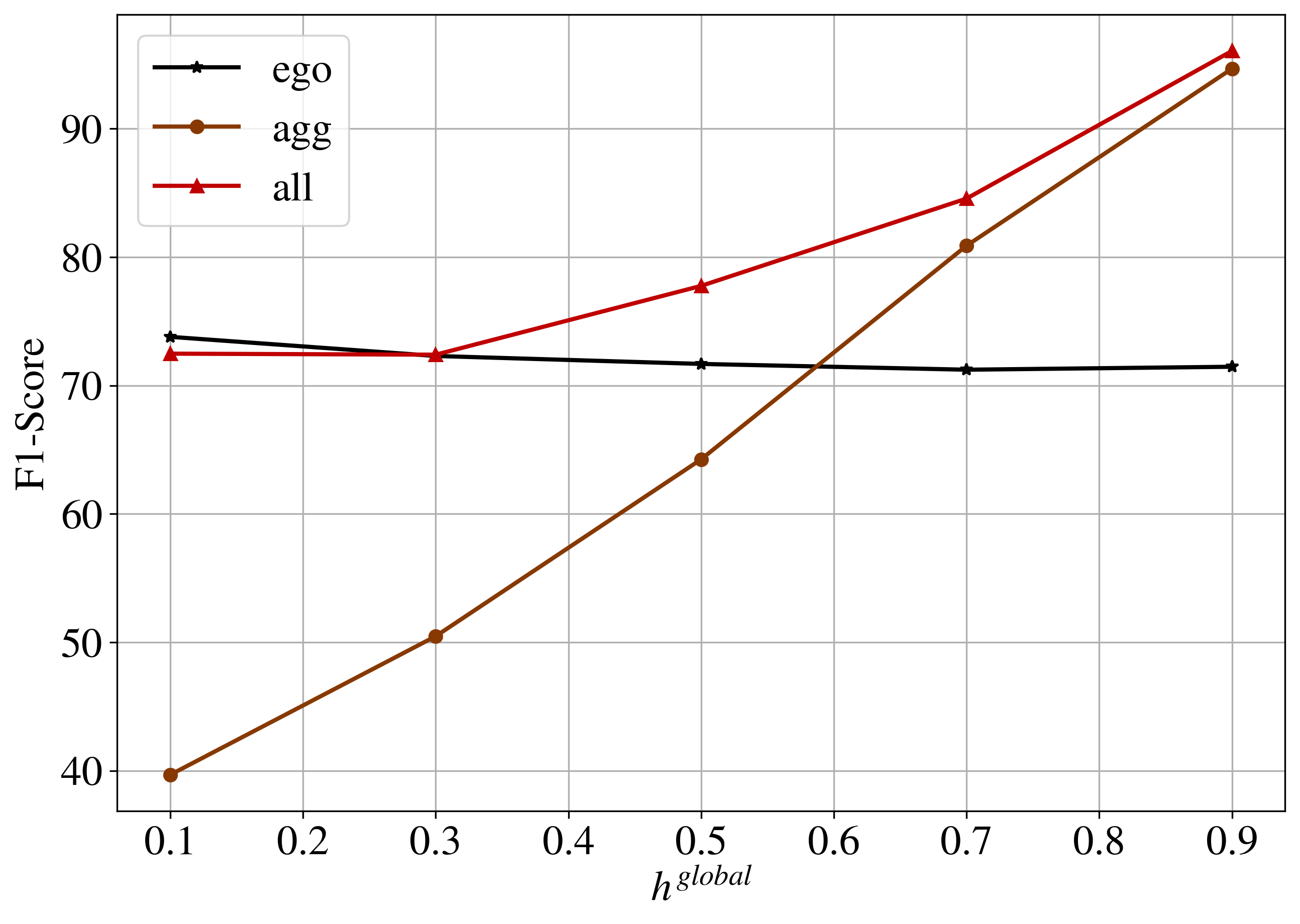}
  \caption{syn-cora}
 \end{subfigure}
 \begin{subfigure}{0.32 \textwidth}
  \centering
  \includegraphics[width=\textwidth]{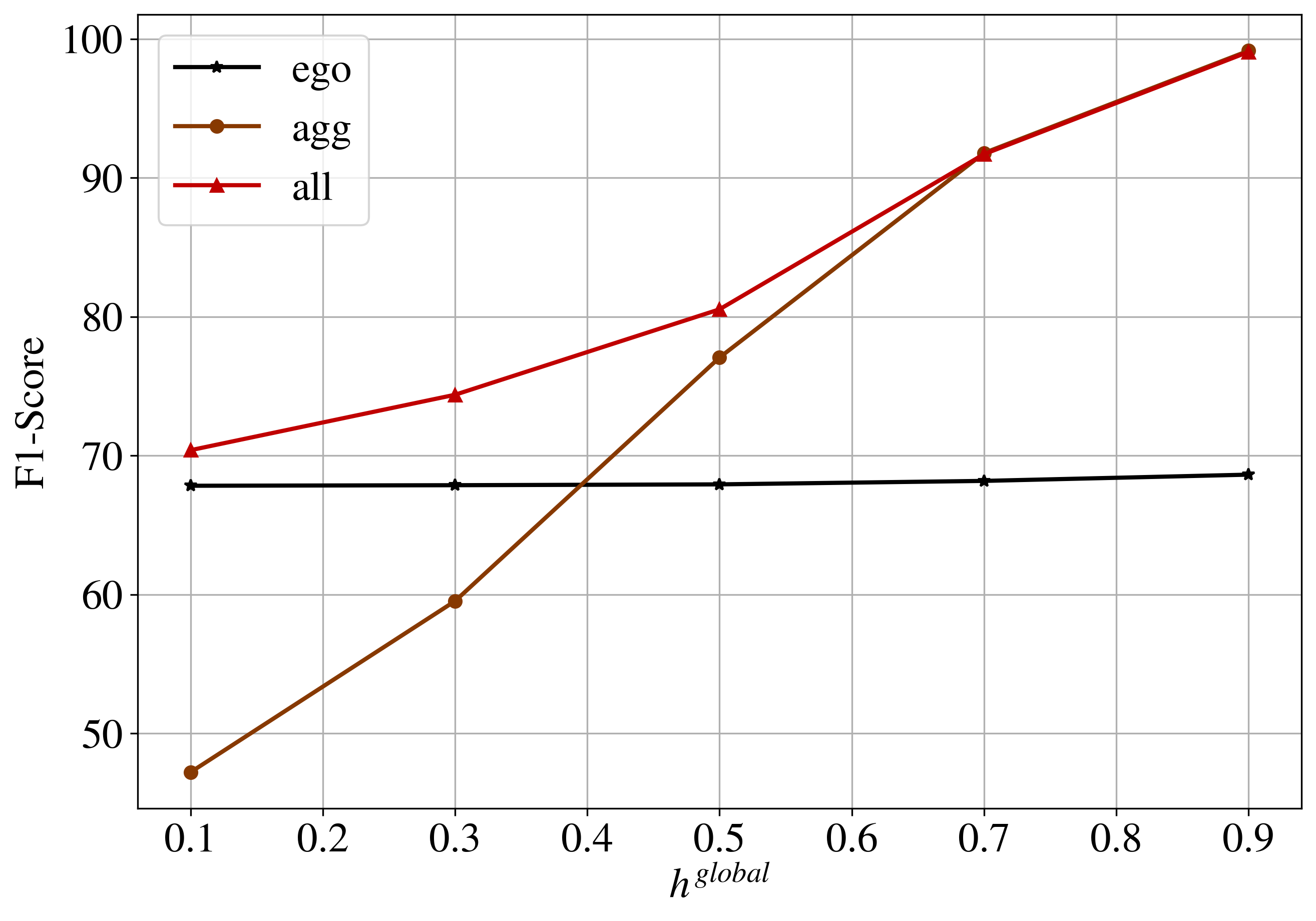}
  \caption{syn-products}
 \end{subfigure}
 \caption{Node classification results on synthetic datasets with different global/local homophily. (a)-(c) shows the classification performance of the existing typical GRL method on syn-cora. (d)-(e) are the ablation analysis of ego task and agg task.
 }
 \label{fig:NodeClassificationOfSyn}
\end{figure}

\noindent\textbf{Performance comparison under different global homophily.} Fig.~\ref{fig:CompOfSyn1} shows the Micro-F1 score of node classification on syn-cora. While all models achieve near-perfect performance under strong homophily, the performance of baselines based on homophily assumptions, e.g., DGI and SAGE-U, significantly deteriorates under heterophily settings. In comparison, MVGE has the best overall performance, outperforming other baselines, especially in heterophily settings, while tying with other models in homophily. PairE, which encodes different signals in the embeddings, has more advantages than other baselines under heterophily settings but still has weak performance.

\noindent\textbf{Performance comparison under different local homophily.} To better analyze why MVGE can achieve good performance in graphs with different homophily settings, we conduct an analytical experiment by evaluating the node classification performance on nodes with different local homophily defined in Eq. \ref{equ:local} on syn-cora with global homophily 0.1 and 0.7. According to local homophily, we divide nodes into three different ranges, [0,0.33), [0.33,0.66) and [0.66, 1]. Fig.~\ref{fig:CompOfSyn2} and ~\ref{fig:CompOfSyn3} show node classification results over different local homophily intervals. We can clearly see that MVGE keeps a rather good performance on the nodes with low local homophily, which are difficult to identify their classes under homophily assumption. In comparison, DGI and SAGE-U suffer poor performance on those difficult nodes, i.e., 25.42\% in $h^{global}=0.1$ and 36.62\% in $h^{global}=0.7$, for nodes with strong heterophily (0$\sim$0.33). They have comparable good performance for nodes with high local homophily (0.66$\sim$1), i.e., 97.05\% and 91.63\%. PairE, which leverages low-frequency and high-frequency signals, learns more effectively under heterophily settings. However, it uses paired node as a basic embedding unit and needs the translation between pair embeddings to node embeddings. It suffers information loss during this translation. MVGE achieves consistent performance on both local heterophily and homophily and outperforms the other three baselines.



\noindent\textbf{Impact of ego-task/agg-task under different global homophily.} To discuss the performance impact of the two feature reconstruction tasks on downstream tasks under different global homophily, we conduct node classification on two synthetic benchmark syn-cora and syn-products. Fig.~ \ref{fig:NodeClassificationOfSyn} shows the classification results at 30$\%$ training ratio. We can observe that the performance under the two feature reconstruction tasks is generally better than the single pretext task. It verifies the importance of integrating different information into node embeddings. On the other hand, comparing the effects of ego-task and agg-task, we can observe that for networks with strong heterophily, embeddings from ego-task generally achieve much better performance, even better than all-task in syn-cora with global homophily 0.1. In this graph, a node significantly differs from its neighbors. In this case,  the agg-task might collect much noise that can perturb the learning process and deviate node representations from the ground truth. In comparison, the ego-task guarantees the basic facts by preserving the high-frequency signals of the nodes themselves. For networks with strong homophily, only the agg-task can achieve good performance. The reason is that the agg-task essentially enhances the low-frequency signals (similar parts) in the neighborhood information. Of course, due to the lack of labels during GRL, we actually do not know the actual trend of connectivity during embedding. Thus, it is important to use two pretext tasks.

\subsection{Results on real-world Datasets}

In this section, we evaluate MVGE with three different downstream tasks. Each of them normally demands different types of information for classification. 

\subsubsection{Results on Node Classification Task} 
\label{ssec:nodeclassification}
We report the overall node classification results in Table \ref{tab:NodeClassification}. We can observe from the table that MVGE outperforms all unsupervised baselines on all eight datasets with different global homophily. For example, in Cornell, MVGE achieves the average Micro-F1 score of 78.89\%, and up to 12\% relative performance improvements over the compared baselines. Experiment results clearly show that MVGE can effectively retain rich signals during embeddings with its multi-view self-supervision design and the distribution-based reconstruction tasks that can keep both low-frequency and high-frequency signals in node features. Other baselines generally achieve good performance in the graphs with homophily while comparably poor performance in heterophily settings. There still exists a large performance gap compared with the models beyond homophily assumption. For example, the contrastive method MVGRL achieves 82.09\% Micro-F1 score in Cora with strong homophily, while only 58.85\% in Cornell with strong heterophily.

\begin{table}[tbp]
  \caption{Node classification results, evaluated with average Micro-F1, \textbf{bold font} represent the best results. "$-$" denotes that the model can not finish the training in 5 days.}
  \label{tab:NodeClassification}
  \resizebox{\textwidth}{!}{
  \begin{tabular}{p{4.5em}cccccccc}
    \hline
    \textbf{Name}&\textbf{Cora}&\textbf{Citeseer}&\textbf{Pubmed}&\textbf{Cora$\_$full}&\textbf{Chameleon}&\textbf{Cornell}&\textbf{Texas}&\textbf{Wisconsin}\\
    \textbf{\small{Hom.ratio}} &\textbf{0.81} &\textbf{0.74} &\textbf{0.8} &\textbf{0.57} &\textbf{0.23} &\textbf{0.3} &\textbf{0.11} &\textbf{0.21}\\
    \hline
    DeepWalk &80.05$\pm$\small{5.74}  &56.02$\pm$\small10.14 &79.80$\pm$\small3.66 &37.44$\pm$\small3.73 &24.16$\pm$\small0.17 &48.67$\pm$\small27.84 &47.00$\pm$\small29.28 &48.67$\pm$\small27.84 \\
    SAGE-U   &80.18$\pm$\small9.85  &71.34$\pm$\small4.32  &83.82$\pm$\small3.01 &55.77$\pm$\small1.54 &42.90$\pm$\small9.27 &48.18$\pm$\small13.61 &61.27$\pm$\small11.93 &60.00$\pm$\small8.43 \\
    DGI-gcn       &81.28$\pm$\small7.58  &68.94$\pm$\small9.10  &83.11$\pm$\small1.50 &43.46$\pm$\small1.66 &42.90$\pm$\small7.33 &38.73$\pm$\small11.93 &77.82$\pm$\small18.31 &60.00$\pm$\small22.31  \\
    GAE      &78.20$\pm$\small4.61  &56.32$\pm$\small7.89  &81.89$\pm$\small1.87 &54.96$\pm$\small1.68 &39.45$\pm$\small5.56 &63.09$\pm$\small9.87  &59.09$\pm$\small16.26 &54.67$\pm$\small21.33 \\
    ARVGE    &79.06$\pm$\small7.94  &60.42$\pm$\small10.01 &71.50$\pm$\small3.18 &45.11$\pm$\small2.21 &41.61$\pm$\small6.11 &54.33$\pm$\small30.27 &50.33$\pm$\small28.20 &54.42$\pm$\small21.84  \\
    P-GNN     &63.60$\pm$\small10.28 &46.92$\pm$\small6.04  &69.08$\pm$\small2.42 &46.12$\pm$\small2.16 &33.85$\pm$\small6.67 &53.89$\pm$\small11.17 &57.80$\pm$\small6.78  &49.33$\pm$\small9.49 \\
    MERIT    &41.80$\pm$\small1.46  &68.13$\pm$\small0.23  &85.05$\pm$\small0.10 &$-$             &26.50$\pm$\small0.54               &30.46$\pm$\small4.79  &66.90$\pm$\small13.09 &54.66$\pm$\small5.33\\
    MVGRL    &82.09$\pm$\small5.25  &62.64$\pm$\small4.41  &80.12$\pm$\small1.71 &52.99$\pm$2.04 &41.19$\pm$\small6.64 &58.85$\pm$\small16.79 &62.22$\pm$\small18.33 &60.13$\pm$\small18.75\\
    PairE    &86.51$\pm$\small8.52  &72.14$\pm$\small2.53  &85.73$\pm$\small2.34 &61.02$\pm$0.95 &45.52$\pm$\small3.37 &66.73$\pm$\small7.77 &61.51$\pm$\small4.86 &68.00$\pm$\small9.98 \\
    \textbf{MVGE} &\textbf{86.95}$\pm$\textbf{\small{2.40}} &\textbf{75.35}$\pm$\small\textbf{0.95} &\textbf{86.82}$\pm$\small\textbf{0.85} &\textbf{66.90}$\pm$\small\textbf{1.98} &\textbf{45.68}$\pm$\small\textbf{1.79} &{\textbf{78.89}$\pm$\small\textbf{9.19}} &{\textbf{81.27}$\pm$\small\textbf{10.05}} &{\textbf{79.33}$\pm$\small\textbf{9.10}} \\
      \hline
    \end{tabular}
}
\end{table}

\begin{table}[tbp]
  \caption{Link prediction task, measured in ROC\_AUC. Standard deviation errors are given. \textbf{bold font} respectively represent the best results in all algorithms. "$-$" is similar to Table~\ref{tab:NodeClassification}}
  \label{tab:LinkPred}
  \resizebox{\textwidth}{!}{
  \begin{tabular}{ccccccccc}
    \toprule
    \textbf{Name}&\textbf{Cora}&\textbf{Citeseer}&\textbf{Pubmed}&\textbf{Cora$\_$full}&\textbf{Chameleon}&\textbf{Cornell}&\textbf{Texas}&\textbf{Wisconsin}\\
    \midrule
  DeepWalk  &74.35$\pm$\small0.83 &77.79$\pm$\small0.53 &72.84$\pm$\small0.71 &86.69$\pm$\small0.27 &74.23$\pm$\small0.59 &55.97$\pm$\small4.28 &55.39$\pm$\small5.41  &66.69$\pm$\small2.94\\
  SAGE-U    &89.97$\pm$\small0.39 &92.44$\pm$\small0.22 &87.97$\pm$\small0.23 &95.73$\pm$\small0.04 &82.64$\pm$\small0.64 &75.98$\pm$\small4.60 &63.91$\pm$\small2.53  &77.21$\pm$\small2.86\\
  DGI-gcn   &83.70$\pm$\small1.00 &85.74$\pm$\small2.24 &84.57$\pm$\small0.33 &79.20$\pm$\small0.56 &82.22$\pm$\small0.44 &82.98$\pm$\small3.52 &83.81$\pm$\small2.90  &84.34$\pm$\small1.99\\
  P-GNN     &82.17$\pm$\small0.04 &77.91$\pm$\small1.44 &82.76$\pm$\small0.17 &90.05$\pm$\small0.15 &87.60$\pm$\small0.80 &48.15$\pm$\small7.08 &48.21$\pm$\small12.37 &64.07$\pm$\small1.11\\
  GAE       &91.18$\pm$\small0.58 &88.00$\pm$\small0.93 &\textbf{96.41}$\pm$\small\textbf{0.11} &96.92$\pm$\small0.10 &\textbf{98.48}$\pm$\small\textbf{0.09} &47.16$\pm$\small3.58 &43.56$\pm$\small2.63  &54.11$\pm$\small3.39\\
  ARVGAE    &92.14$\pm$\small0.24 &89.08$\pm$\small0.47 &92.04$\pm$\small1.21 &95.82$\pm$\small0.61 &97.82$\pm$\small0.16 &56.70$\pm$\small3.12 &48.96$\pm$\small2.56  &75.38$\pm$\small1.62\\
  MERIT     &85.84$\pm$\small0.10 &61.95$\pm$\small0.32 &63.90$\pm$\small0.23 &$-$ &52.44$\pm$\small1.38 &70.11$\pm$\small1.36 &56.25$\pm$\small5.86  &53.64$\pm$\small3.30\\
  MVGRL     &84.52$\pm$\small1.47 &60.38$\pm$\small2.23 &80.91$\pm$\small1.35 &78.59$\pm$\small0.96 &80.72$\pm$\small0.55 &75.05$\pm$\small5.84 &81.17$\pm$\small6.50 &76.08$\pm$\small6.95\\
  PairE     &92.92$\pm$\small0.41 &95.10$\pm$\small0.20 &94.98$\pm$\small0.23 &96.91$\pm$\small0.09 &93.63$\pm$\small0.27 &80.57$\pm$\small2.47 &83.89$\pm$\small1.56 &86.47$\pm$\small2.64\\
  MVGE      &\textbf{94.86}$\pm$\small\textbf{0.31} &\textbf{95.36}$\pm$\small\textbf{0.16} &96.20$\pm$\small0.06 &\textbf{97.72}$\pm$\small\textbf{0.04} &94.76$\pm$\small0.08  &\textbf{85.15}$\pm$\small\textbf{1.46} &\textbf{84.58}$\pm$\small\textbf{1.38} &\textbf{90.33}$\pm$\small\textbf{1.50}\\
  \bottomrule
\end{tabular}
}
\end{table}

\begin{table}[tbp]
  \caption{pair-wise prediction task, measured in ROC\_AUC. Standard deviation errors are given. \textbf{bold font} respectively represent the best results in all algorithms. "$-$" is similar to Table~\ref{tab:NodeClassification}}
  \label{tab:PairPred}
  \resizebox{\textwidth}{!}{
  \begin{tabular}{ccccccccc}
    \toprule
    \textbf{Name}&\textbf{Cora}&\textbf{Citeseer}&\textbf{Pubmed}&\textbf{Cora$\_$full}&\textbf{Chameleon}&\textbf{Cornell}&\textbf{Texas}&\textbf{Wisconsin}\\
    \midrule
    DeepWalk &60.37$\pm$\small1.24 &53.05$\pm$\small1.47 &53.05$\pm$\small0.27 &61.91$\pm$\small0.51 &58.20$\pm$\small0.61 &52.58$\pm$\small3.19 &60.57$\pm$\small3.95 &52.50$\pm$\small1.78\\
    SAGE-U   &57.17$\pm$\small1.53 &50.85$\pm$\small1.65 &50.47$\pm$\small0.18 &53.34$\pm$\small0.59 &50.18$\pm$\small0.61 &70.49$\pm$\small4.07 &71.73$\pm$\small4.30 &71.23$\pm$\small1.69\\
   DGI-gcn   &62.81$\pm$\small1.33 &51.99$\pm$\small1.42 &58.43$\pm$\small1.82 &53.71$\pm$\small1.58 &49.72$\pm$\small0.40 &65.45$\pm$\small3.88 &71.44$\pm$\small4.39 &59.88$\pm$\small1.52\\
   GAE       &75.84$\pm$\small2.87 &65.09$\pm$\small1.14 &76.28$\pm$\small1.00 &86.91$\pm$\small0.06 &57.78$\pm$\small0.86 &53.51$\pm$\small2.30 &56.23$\pm$\small4.24 &52.49$\pm$\small5.16\\
   ARVGAE    &77.38$\pm$\small1.43 &64.75$\pm$\small1.77 &73.28$\pm$\small1.11 &85.91$\pm$\small0.56 &60.90$\pm$\small0.30 &53.62$\pm$\small4.98 &57.99$\pm$\small8.63 &57.63$\pm$\small4.53\\
   P-GNN     &66.40$\pm$\small0.83 &58.07$\pm$\small0.98 &61.69$\pm$\small0.20 &74.18$\pm$\small0.32 &58.47$\pm$\small0.55 &51.97$\pm$\small4.49 &52.73$\pm$\small3.53 &49.32$\pm$\small1.88\\
   MERIT     &67.08$\pm$\small1.12 &53.67$\pm$\small0.81 &60.23$\pm$\small0.31 &$-$ &51.04$\pm$\small0.25 &72.74$\pm$\small1.28 &69.7$\pm$\small1.46 &63.25$\pm$\small3.87\\
   MVGRL     &81.49$\pm$\small0.58 &64.03$\pm$\small0.64 &81.62$\pm$\small1.00 &90.10$\pm$\small0.20 &57.62$\pm$\small1.02 &62.47$\pm$\small6.58 &71.17$\pm$\small1.13 &59.95$\pm$\small5.58\\
   PairE &66.54$\pm$\small1.46 &60.95$\pm$\small0.96 &69.67$\pm$\small0.63 &73.64$\pm$\small0.72 &58.64$\pm$\small0.67 &72.62$\pm$\small3.51 &71.89$\pm$\small4.40 &73.56$\pm$\small2.95\\
   MVGE      &\textbf{83.25}$\pm$\small\textbf{0.49} &\textbf{79.61}$\pm$\small\textbf{0.56} &\textbf{84.13}$\pm$\small\textbf{0.03} &\textbf{90.21}$\pm$\small\textbf{0.11} &\textbf{60.97}$\pm$\small\textbf{0.29} &\textbf{87.89}$\pm$\small\textbf{0.46} &\textbf{85.61}$\pm$\small\textbf{0.53} &\textbf{87.85}$\pm$\small\textbf{1.54}\\
  \bottomrule
\end{tabular}
}
\end{table}



\subsubsection{Results on Link Prediction task}
\label{ssec:linkprediction}

Table \ref{tab:LinkPred} shows the ROC\_AUC scores of the link prediction task on all eight datasets. We can find that our model achieves state-of-the-art results with respect to unsupervised models in six out of eight datasets. For example, in Wisconsin, we achieve a 90.33\%  ROC\_AUC score which is almost 4\% relative improvement over the previous state-of-the-art. It attributes to the fact that we design an adjacency matrix reconstruction task to predict whether there is a link between two nodes directly. In addition, the random walk operation in data augmentation also captures the structural information of different local neighborhoods in the network to a certain extent. 

GAE and ARVGE, which adopt link reconstruction in the decoder, also achieve competitive performance under homophily settings. However, their homophily assumption making connected nodes with similar embeddings, these models fail to predict links accurately in heterophily settings, e.g., Cornell, Texas, and Wisconsin.



\subsubsection{Results on Pair-wise Prediction task}
Table \ref{tab:PairPred} summarizes the performance of MVGE and other baselines on pairwise node classification tasks, where MVGE gains the best classification accuracy on all eight datasets. Specifically, MVGE achieves 89.89\% ROC\_AUC score while the previous state-of-the-art method MERIT can only achieve 72.74\% ROC\_AUC, which means that MVGE gives about 15\% relative improvement on this task. 

In the pairwise node classification task, we believe that the implicit relationship of the pair is related to the low- and high-frequency information of the features of the paired nodes. In other words, two nodes with similar embeddings are more likely to fall into the same class. For paired nodes belonging to different classes, the embeddings of paired nodes should contain enough high-frequency information for classification. Results show that MVGE has a superior classification performance and can effectively differentiate nodes from different classes. While models based on homophily assumption are concerned with capturing the commonality of node features and inevitably ignore the difference. Their designs make rich relationship patterns become vague and indistinguishable in learned representation. Thus, their performance in graphs with heterophily is rather poor, e.g., DGI 59.88\% and MGRL 59.95\% v.s. MVGE 87.85\%.



\begin{figure}[htbp]
\centering
\begin{subfigure}{0.24\textwidth}
  \centering
  \includegraphics[width=\textwidth]{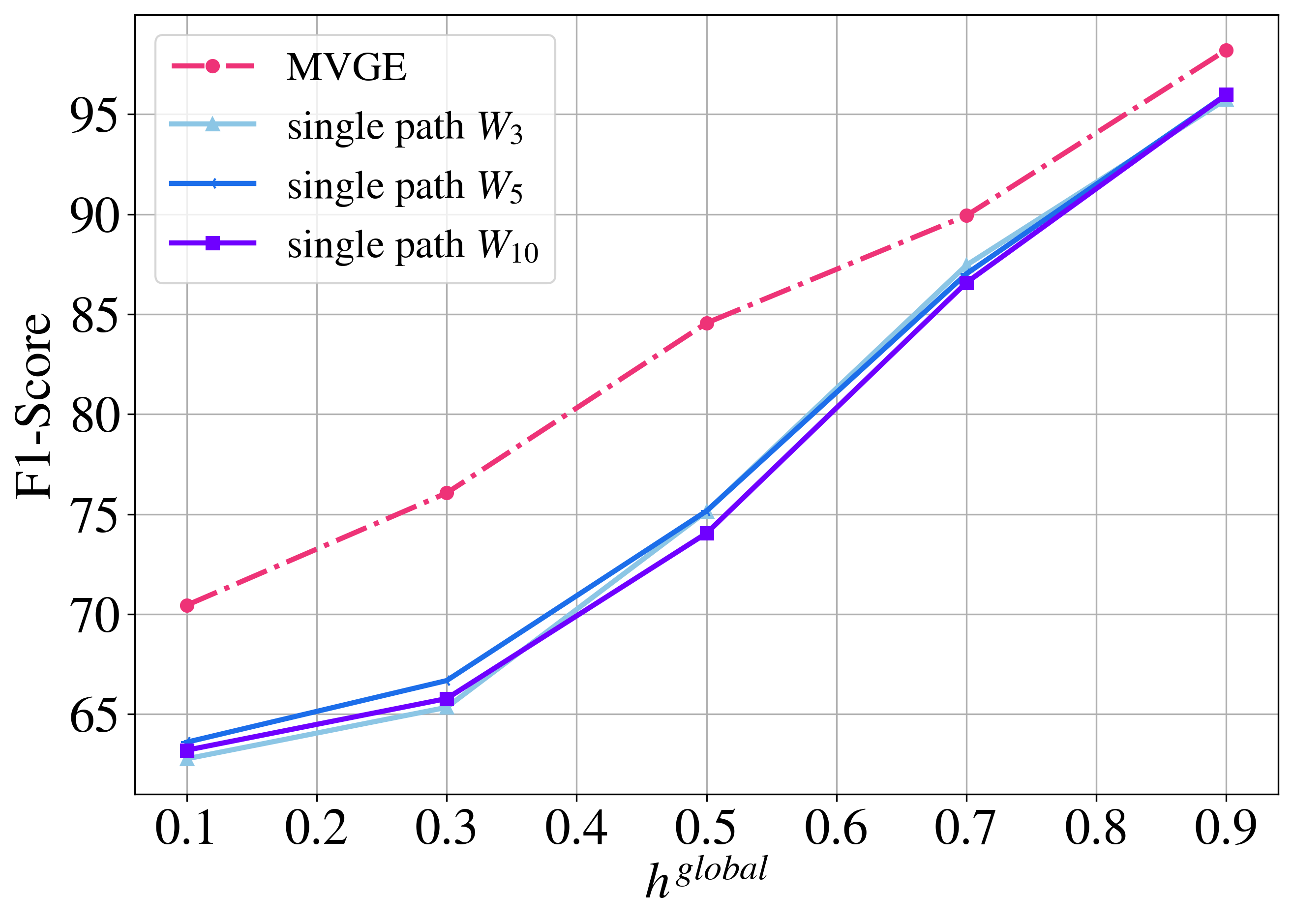}
  \caption{syn-cora}
  \label{fig:deepwalk1}
 \end{subfigure}
 \begin{subfigure}{0.24\textwidth}
  \centering
  \includegraphics[width=\textwidth]{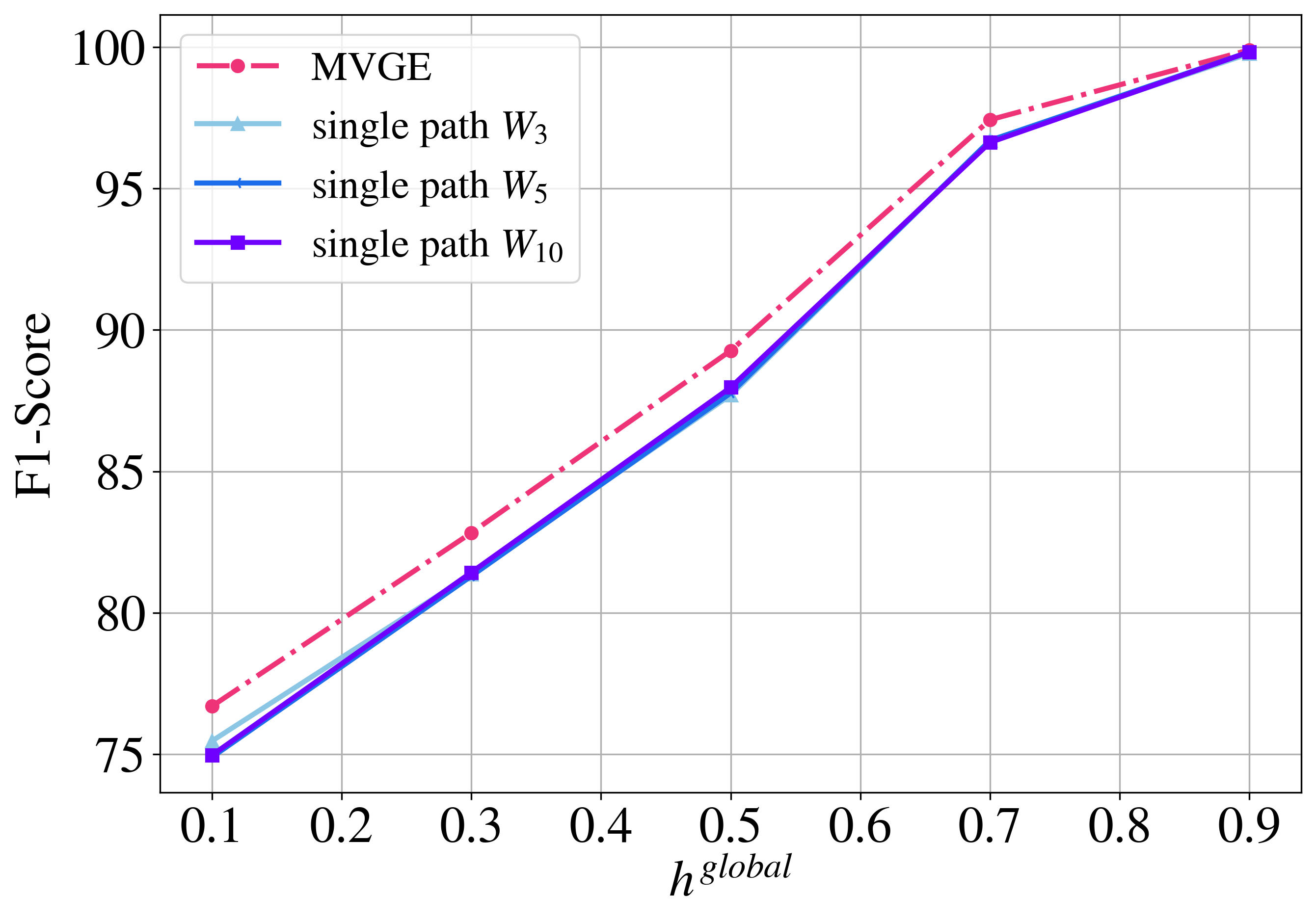}
  \caption{syn-products}
  \label{fig:deepwalk2}
 \end{subfigure}
 \begin{subfigure}{0.24\textwidth}
  \centering
  \includegraphics[width=\textwidth]{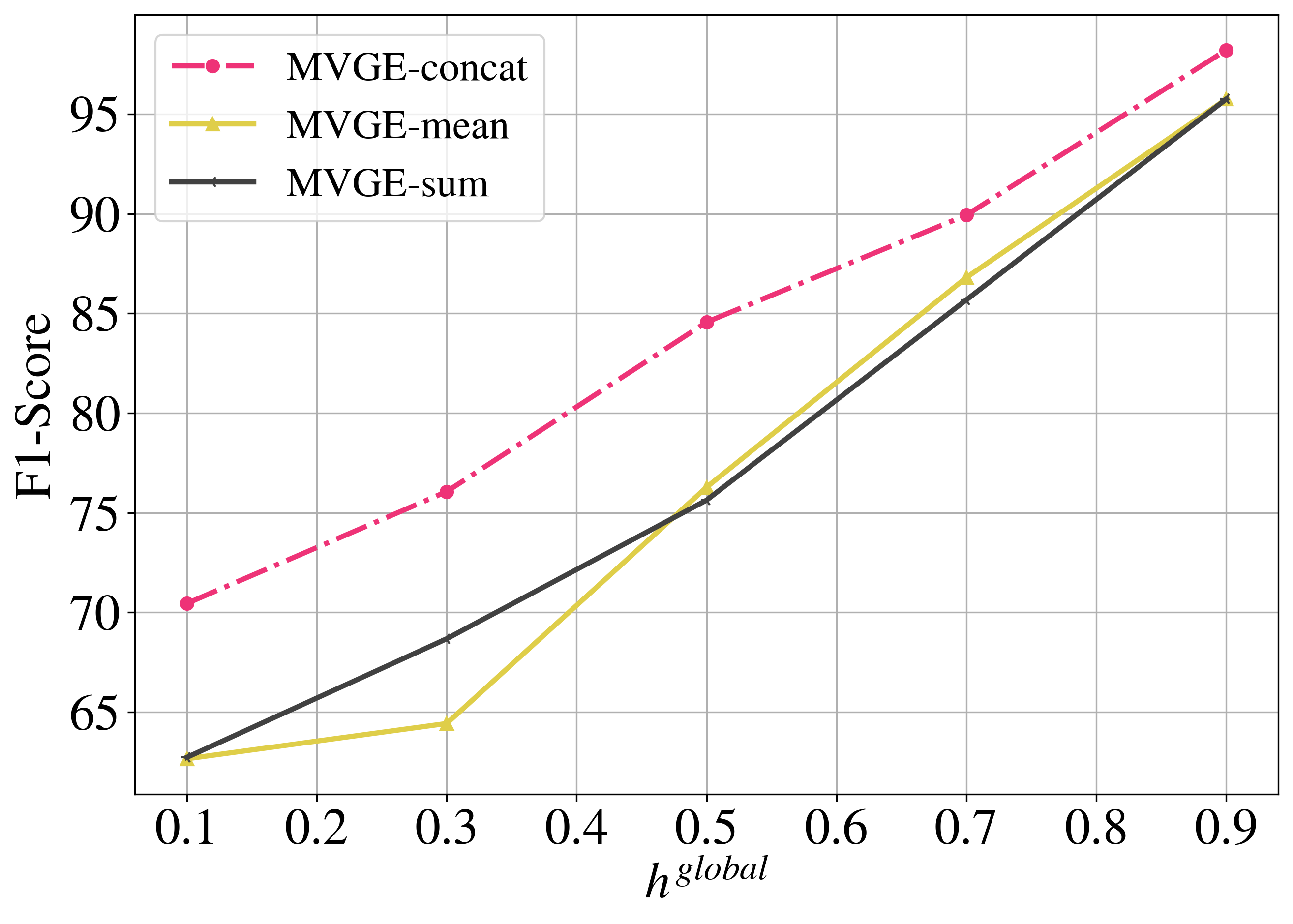}
  \caption{syn-cora}
  \label{fig:deepwalk3}
 \end{subfigure}
  \begin{subfigure}{0.24\textwidth}
  \centering
  \includegraphics[width=\textwidth]{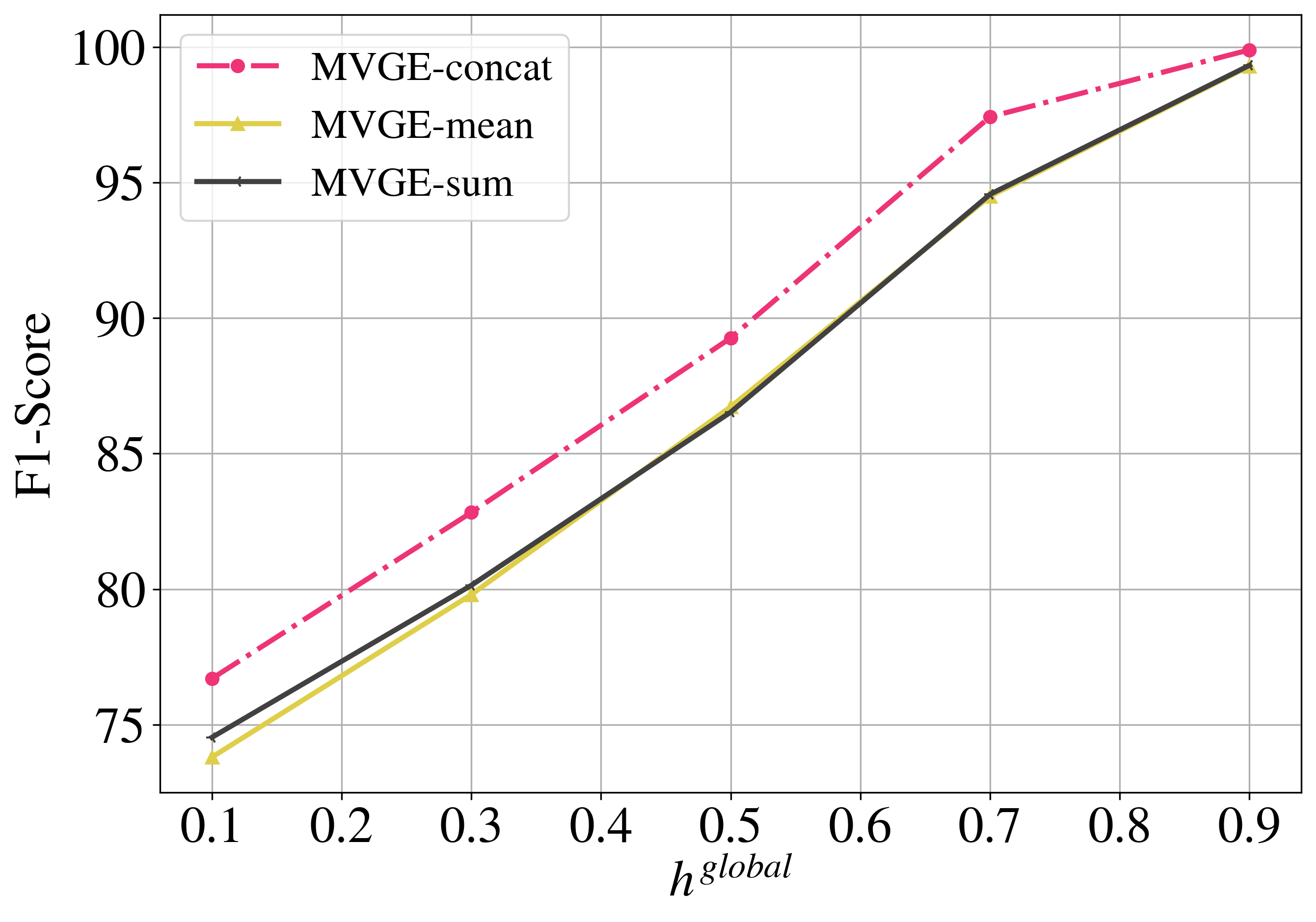}
  \caption{syn-products}
  \label{fig:deepwalk4}
 \end{subfigure}
 \caption{(a)-(b): performance of MVGE for different number of random walk path. (c)-(d): performance of MVGE for different aggregated function.}
 \label{fig:ablation_deepwalk}
\end{figure}

\subsection{Ablation Study}
In this section, we analyze and verify the effectiveness of different designs.

\subsubsection{Design choices for walk-based aggregated features.} 
This subsection analyzes and evaluates the contribution of multiple random walk paths and impacts of different aggregators in Definition 4 based on node classification on synthetic datasets. (1) Fig.~\ref{fig:deepwalk1} and \ref{fig:deepwalk2} shows the results for MVGE and three variants, where $W_3$, $W_5$, and $W_{10}$ denotes there is only single walk path in Eq. (\ref{eq:walk_agg}) with different length respectively.  We observe that MVGE consistently performs better than all the variants, indicating that the agg features obtained by combining multiple walk paths contain the richest information. (2) Furthermore, we consider compare the performance of MVGE with different $AGGR$ function in Eq. (\ref{eq:walk_agg}), i.e., $concat(\cdot)$, $mean(\cdot)$, and $sum(\cdot)$. Fig.~\ref{fig:deepwalk3} and \ref{fig:deepwalk4} show that the $concat(\cdot)$ generally achieves that best performance in all global homophily settings and significantly outperforms the others ($mean(\cdot)$) on syn-cora ($h^{global} = 0.3$) by up to 13\%. These results support the effectiveness of the concatenation operation.


\subsubsection{Effectiveness of Linear Encoder For Ego Features}

\begin{figure}[tbp]
\centering
\begin{subfigure}{0.49\textwidth}
  \centering
  \includegraphics[width=0.75\textwidth]{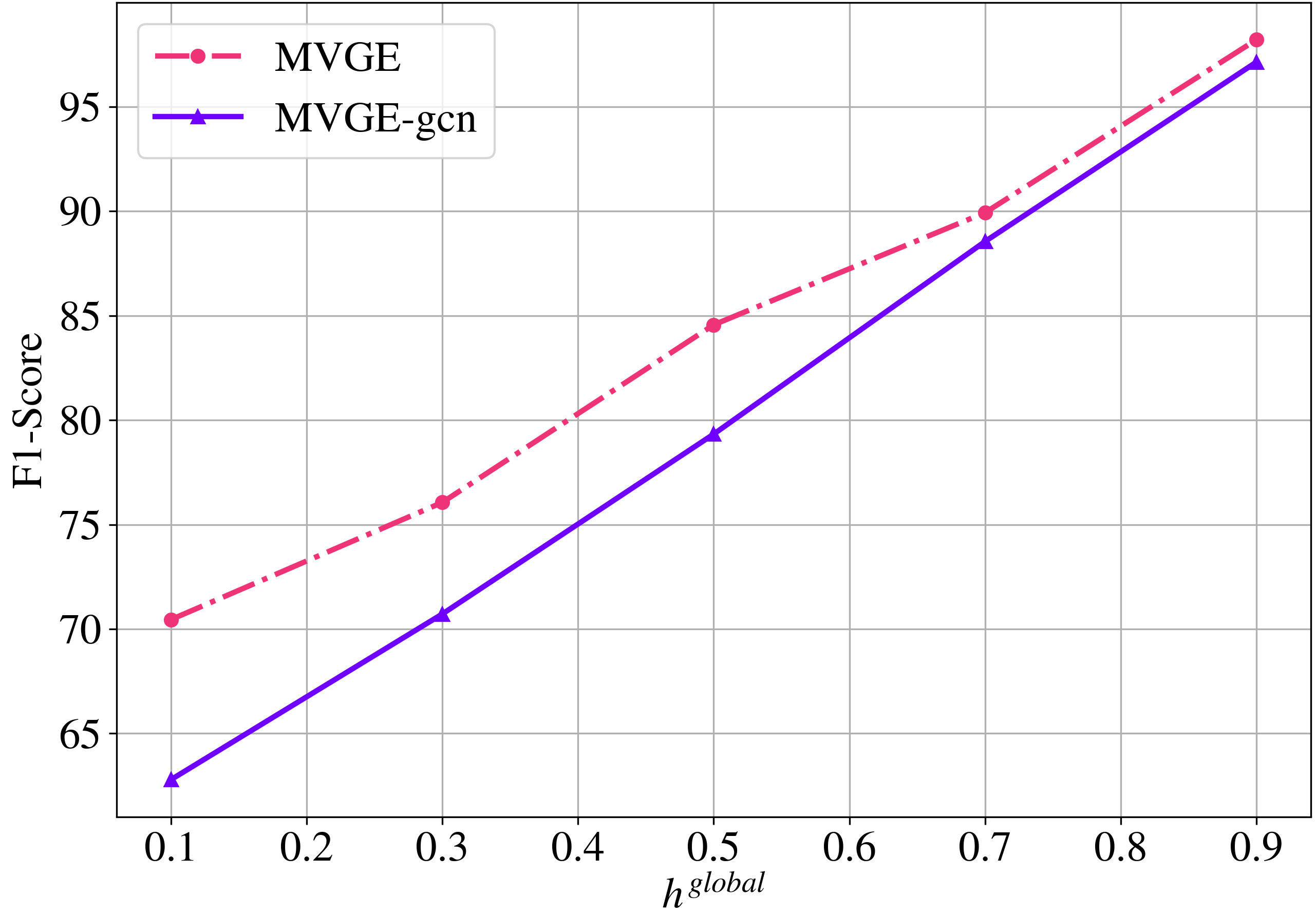}
  \caption{syn-cora.}
 \end{subfigure}
 \begin{subfigure}{0.49\textwidth}
  \centering
  \includegraphics[width=0.75\textwidth]{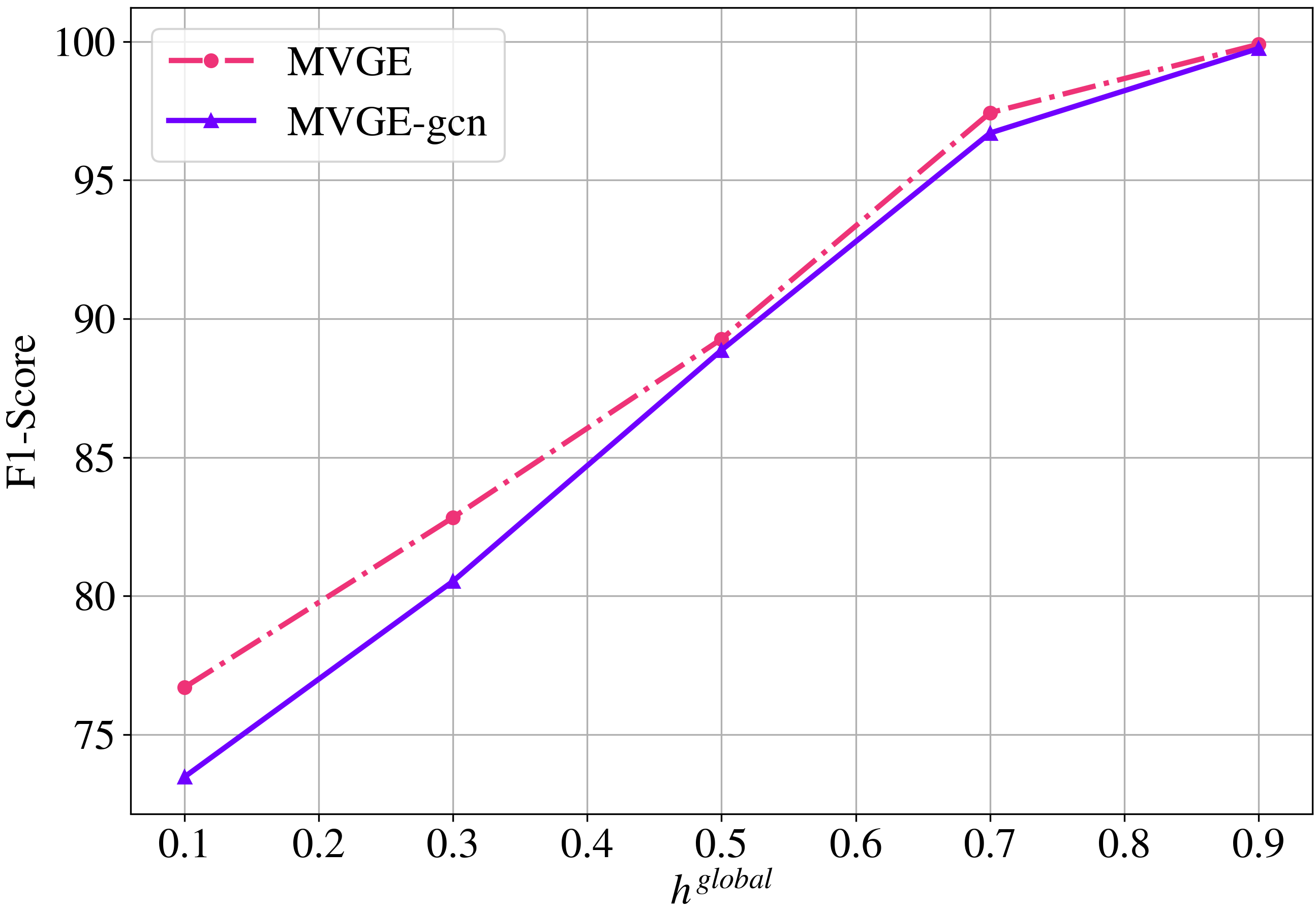}
  \caption{syn-products}
 \end{subfigure}
 \caption{ Node classification performance with different encoders for ego-features on synthetic datasets. 'MVGE-gcn' use two separated GCN encoders. }
 \label{fig:two-gcn}
\end{figure}

To verify the effectiveness of the linear encoder for ego features, we compare it with a variant of MVGE with two separated GCN encoders. Fig.~\ref{fig:two-gcn} shows the node classification results in Micro-F1 score on syn-cora and syn-products with different global homophily. We observe that MVGE with linear encoder consistently performs better than the variant, existing a significantly bigger performance gap in strong heterophily settings. As a popular encoding scheme, the essence of GCN is a low-pass filter: each node repeatedly averages its own features and those of its neighbors to update its own feature representation\cite{nt2019revisiting,xu2020graph}. This mechanism results in final embeddings that are similar across neighboring nodes for any set of original features~\cite{rossi2020proximity}. While this may work well in the case of homophily, where neighbors likely belong to the same class or have similar features, it poses severe challenges in the case of heterophily: it is not possible to distinguish neighbors from different labels based on the similar learned representations. In contrast, a simple linear encoding scheme for ego features can avoid smoothing high-frequency information in features and allow learning distinguishable representations.

\begin{figure}[tbp]
\centering
\begin{subfigure}{0.24\textwidth}
  \centering
  \includegraphics[width=\textwidth]{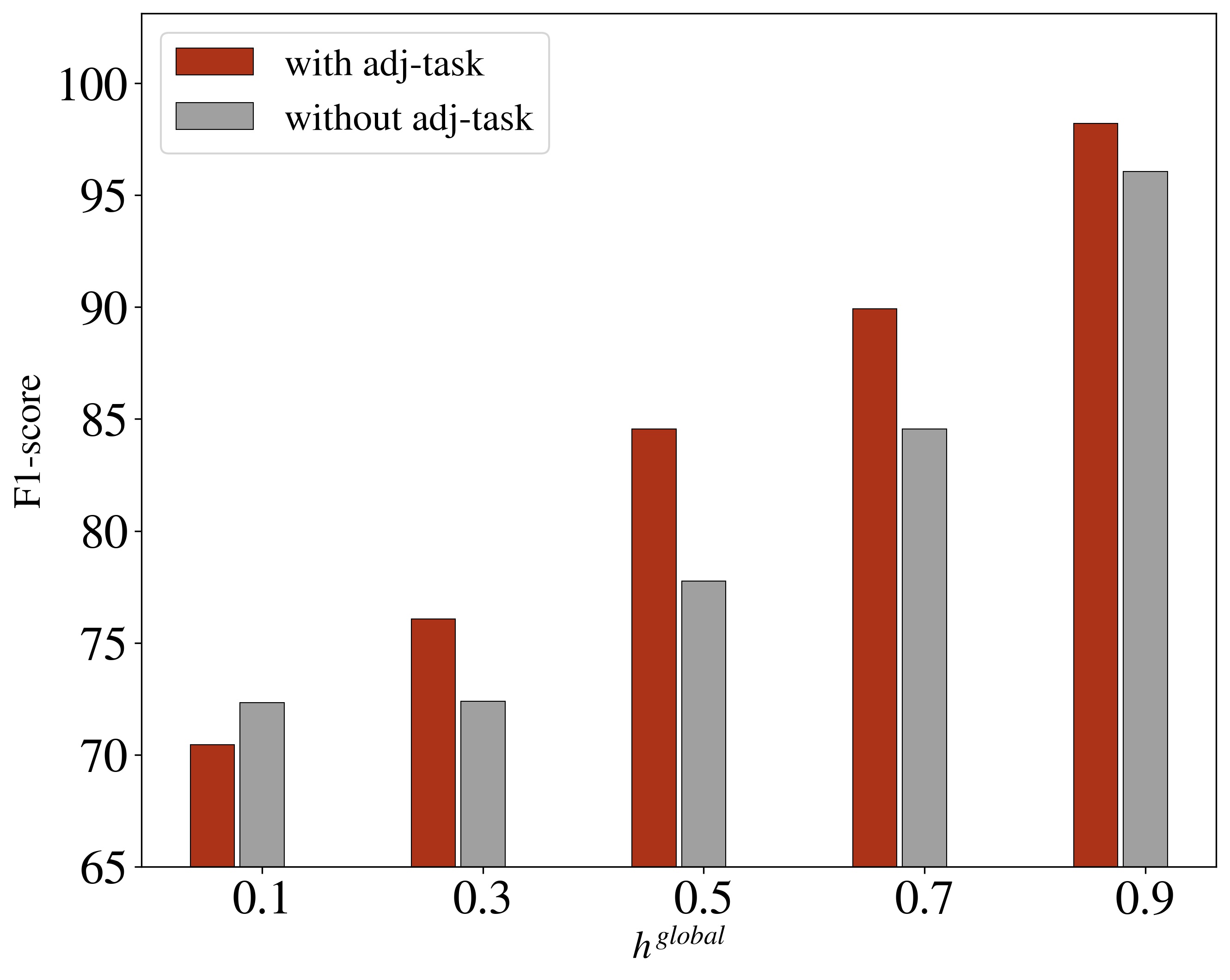}
  \caption{Node. on syn-cora.}
  \label{fig:adj-task1}
 \end{subfigure}
 \begin{subfigure}{0.24\textwidth}
  \centering
  \includegraphics[width=\textwidth]{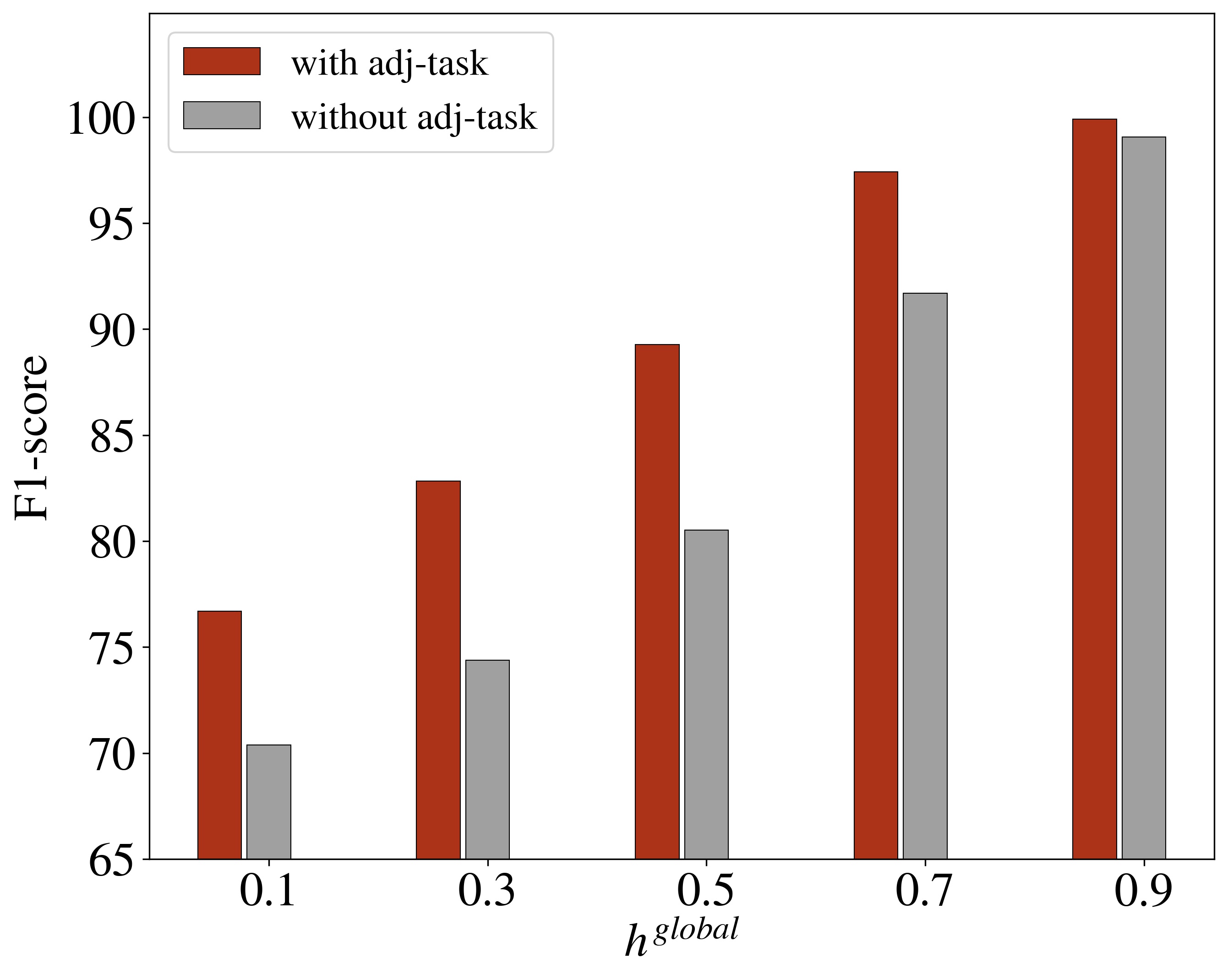}
  \caption{Node. on syn-products}
  \label{fig:adj-task2}
 \end{subfigure}
 \begin{subfigure}{0.24\textwidth}
  \centering
  \includegraphics[width=\textwidth]{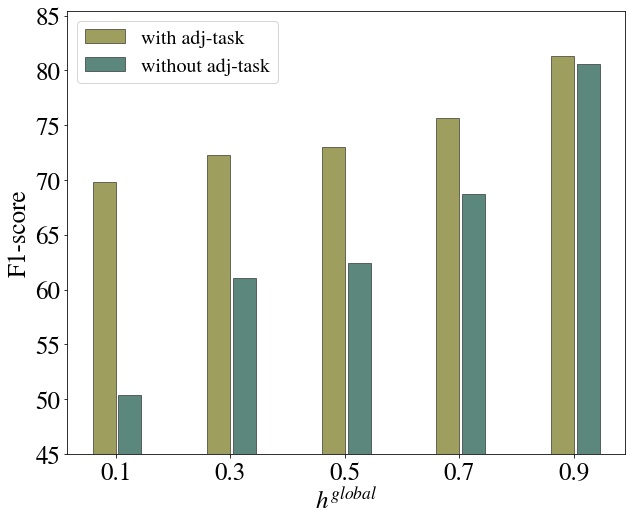}
  \caption{Link. on syn-cora.}
  \label{fig:adj-task3}
 \end{subfigure}
 \begin{subfigure}{0.24\textwidth}
  \centering
  \includegraphics[width=\textwidth]{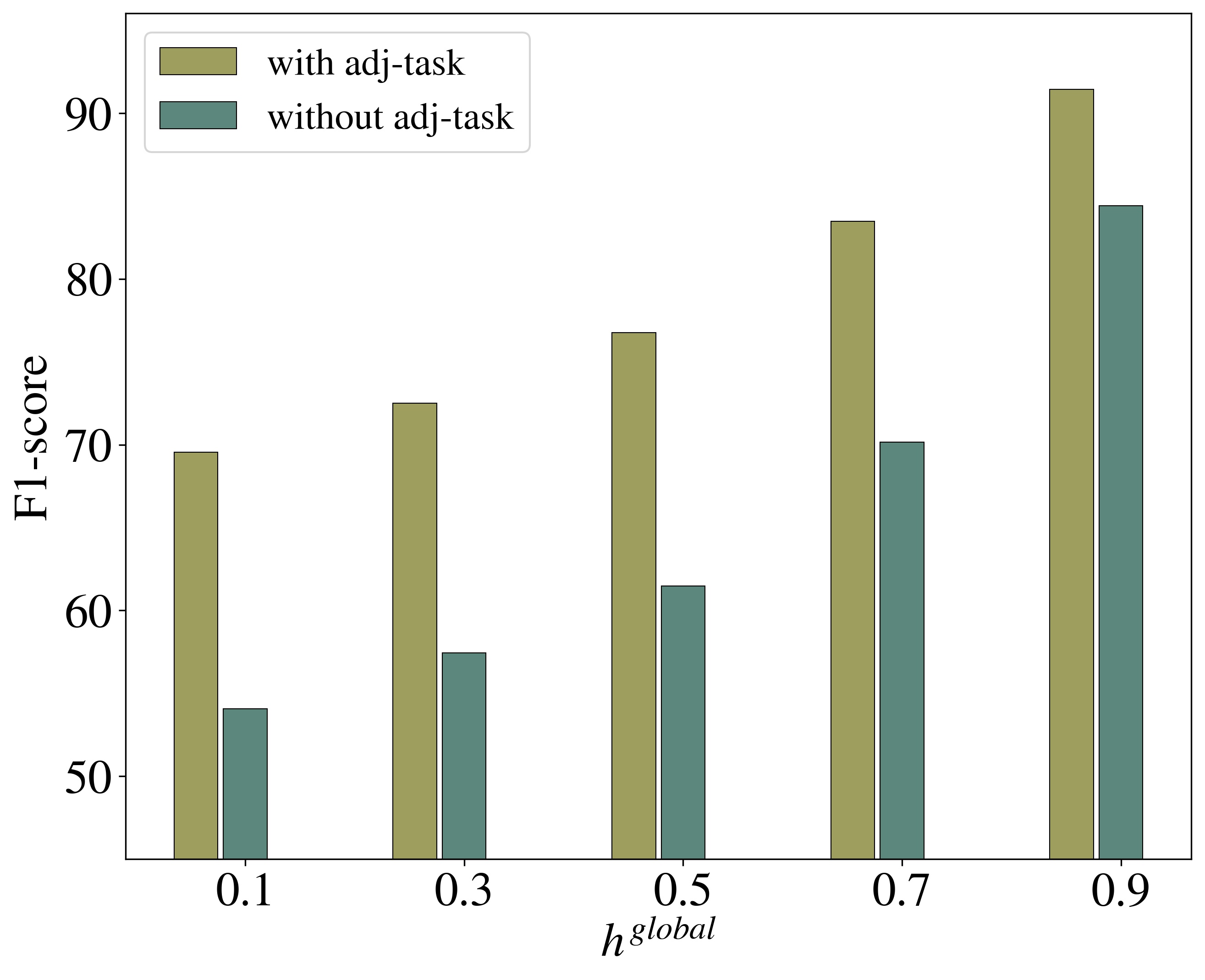}
  \caption{Link. on syn-products.}
  \label{fig:adj-task4}
 \end{subfigure}
 \caption{Effect of the adj-task. (a)/(b) Node classification results on syn-cora and syn-products. (c)/(d) Link prediction results on syn-cora and syn-products.}
 \label{fig:adj-task}
\end{figure}

\subsubsection{Effectiveness of Adjacency Matrix Reconstruction Task.} 
Fig.~\ref{fig:adj-task} gives the node classification and link prediction results of MVGE and a variant without the adjacency reconstruction task on two synthetic datasets. We observe similar trends on both two benchmarks and two downstream tasks: MVGE with adj-task almost has the best overall performance outperforming the variants without adj-task. Results clearly show the importance of merging embeddings with different semantic meanings. In addition, when the global homophily is rather high (0.9), ego embeddings and agg embeddings have very close semantics, which results in a comparably small performance gap between MVGE and its variant. In comparison, when with low global homophily, their performance gaps are generally large. Particularly, as shown in Fig.\ref{fig:adj-task1}, since the purpose of adj-ask is to learn neighboring patterns between connected nodes, e.g., close representations, MVGE with adj-task is less effective than MVGE without adj-task in node classification under strong heterophily (0.1). Furthermore, as the adj-task is able to capture some important characteristics of the graph structure by reconstructing an adjacency matrix $\hat{A}$ close to the true one starting from the node embeddings, the performance of link prediction is greatly improved.

An interesting observation worth noting is that the performance gaps to be opposite in syn-cora and syn-products under strong heterophily ($h^{global}$=0.1). The three core designs involved in this article all have their unique roles: t1, Linear Encoder for Ego-features, to preserve high-frequency information in ego features; t2, GNN Encoder for Agg-features, to learn the commonality among neighbors; t3, Reconstruct the Adjacency Matrix, to learn the proximity of topological distances. t1 and t2 directly preserve the difference and commonality between node features, and their role is to make the distance between the node and its neighbors as consistent as possible in the original feature space and embedding space. While t3 only considers structural information, it is expected that structurally adjacent nodes will also maintain a relatively close distance in the embedding space. 

Therefore, in some cases, the goals of the two will conflict, especially when there are sparse networks with high heterogeneity and low degree (such as syn-cora with heterophily=0.1 and degree = 3.9), at this time, the labels of the central node and its neighbors are all different with a high probability. In such a case, the t1/t2 task will maintain the characteristics of the node $u$ and the commonality with the neighborhood $N(u)$, which usually keeps the embeddings of $u$ and $N(u)$ highly distinguishable, while t3 will make the embeddings of $u$ and $N(u)$ are as similar as possible. At this time, there is a serious conflict between the learning goals of the two, which may cause confusion in learning. For syn-product, its degree is 11, so even if homophily=0.1, there may still be some similar nodes around the central node $u$. Therefore, t1/t2 and t3 can better find consistent signals, and their collaborative learning can also effectively improve performance.  We think this is what causes the performance gaps to be opposite in syn-cora and syn-products under strong heterophily ($h$=0.1). All in all, the impact of homophily level is closely related to the sparsity of the network (node degree), in heterophily, low-degree nodes pose stronger  challenges in learning~\cite{zhu2020beyond}.

\begin{figure}[htbp]
  \centering
  \includegraphics[width=0.6\textwidth]{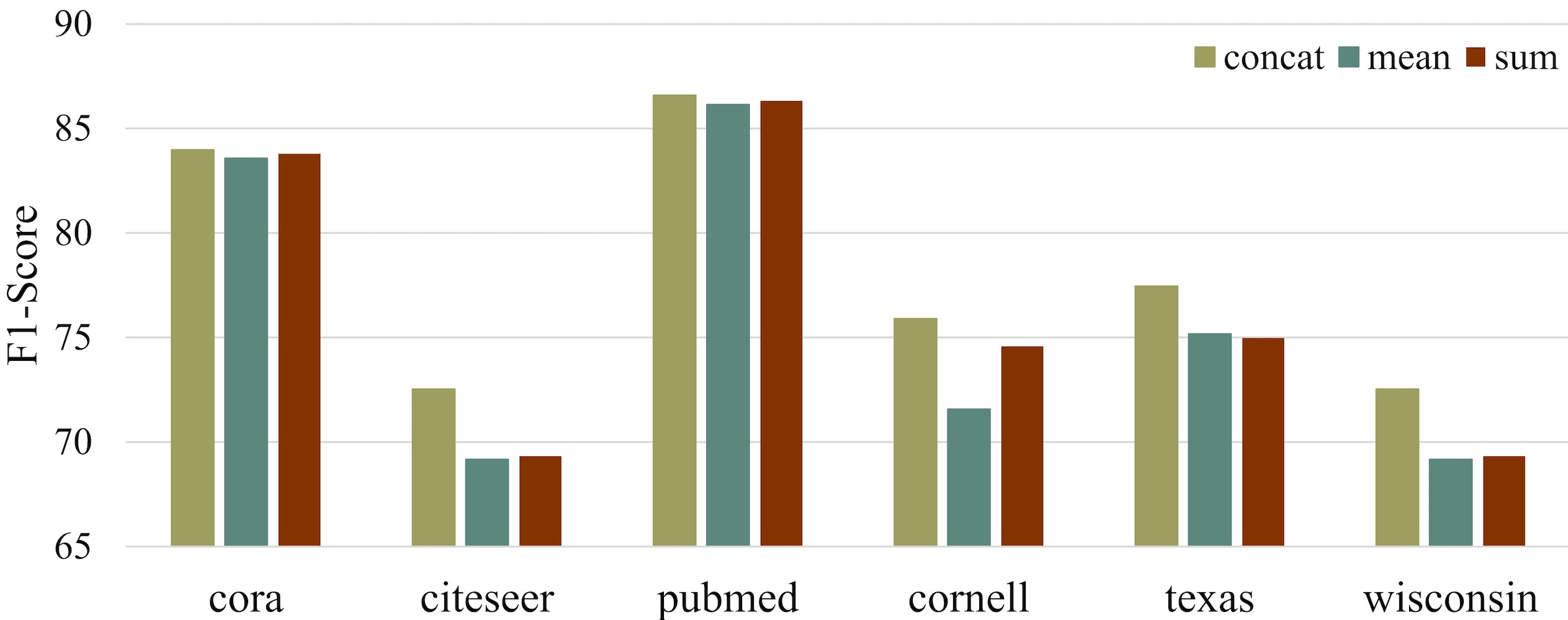}
  \caption{Cp. between different functions in merging ego and agg embeddings on six real-world datasets.}
  \label{fig:Aggregator}
\end{figure}
\subsubsection{Functions in merging ego and agg embeddings}
\label{sec:aggregator}
To validate the effectiveness of the concatenation operation for learned representations, we conduct experiments on six datasets for three MVGE variants with different functions to merge the ego embeddings and agg embeddings, i.e., $concat(\cdot)$,  $sum(\cdot)$ and  $mean(\cdot)$. Fig.~\ref{fig:Aggregator} shows the node classification results under 30\% training ratio. We can observe that $concat\cdot$ function achieves the best performance in all six datasets under different global homophily, as $concat(\cdot)$ function can keep the integrity and personalization of the input information as much as possible, and avoid losses caused by information mixing. The performance of the other two aggregator functions is relatively unstable. Specifically, the $sum(\cdot)$ function generally achieves better performance than the $mean(\cdot)$ function. The embedding accumulation operation of $sum(\cdot)$ can, to a certain extent, preserve the degree information of nodes.

\section{Conclusion}

Due to a lack of support from labels, graph representation learning methods normally make the homophily assumptions on the graph, which result in their poor performance in heterophily settings. Furthermore, one pretext task can hardly capture diverse graph information into task-agnostic embeddings without label supervision. In this paper, we present a novel multi-view graph representation framework MVGE with the usage of diverse pretext tasks to capture different signals in graphs into embeddings. More specifically, a set of new pretext tasks are designed to encode different types of signals, and a straightforward operation is proposed to maintain both the commodity and personalization in both the attribute and the structural levels. Extensive experiments on both synthetic and real-world network datasets show that the node representations learned with MVGE achieve significant performance improvements in different downstream tasks, especially on graphs with heterophily.

\section{Acknowledgements}
This work was supported by the Science and Technology Research and Development Program Project of China railway group limited (Project Nos. 2020-Special-02 and 2021-Special-08). The computing resources supporting this work were partially provided by High-Flyer AI. (Hangzhou High-Flyer AI Fundamental Research Co., Ltd.)


\bibliographystyle{ACM-Reference-Format}
\bibliography{MVGE}

\appendix

\end{document}